\documentclass{article}

\usepackage[preprint]{neurips_2025}

\usepackage[utf8]{inputenc} 
\usepackage[T1]{fontenc}    
\usepackage{hyperref}       
\usepackage{url}            
\usepackage{booktabs}       
\usepackage{amsfonts}       
\usepackage{nicefrac}       
\usepackage{microtype}      
\usepackage{xcolor}         

\usepackage{multirow}
\usepackage{multicol}
\usepackage{amssymb}
\usepackage{colortbl}
\usepackage{longtable}
\usepackage{mdframed}
\usepackage{inconsolata}
\usepackage{color}
\usepackage{amsmath}
\usepackage{todonotes}
\usepackage{enumitem}
\usepackage{makecell}
\usepackage{caption}
\usepackage{subcaption}
\usepackage{pdfpages}
\usepackage{graphicx}
\usepackage{placeins}
\usepackage{pgffor}
\usepackage{algorithm}
\usepackage{algorithmic}

\newcommand{\imp}[1]{{\small\hspace{0.05cm}{\color[HTML]{32CB00}{$_{\textbf{+#1}}$}}}}

\newcommand\footnoteONLYtext[1]{
    \let \mybackup \thefootnote
    \let \thefootnote \relax
    \footnotetext{#1}
    \let \thefootnote \mybackup
    \let \mybackup \imareallyundefinedcommand}

\title{Citrus-V: Advancing Medical Foundation Models with Unified Medical Image Grounding for Clinical Reasoning}

\author{
  \textbf{Guoxin Wang}{\footnotesize $^\bigstar$} \quad
  \textbf{Jun Zhao} \quad
  \textbf{Xinyi Liu} \quad
  \textbf{Yanbo Liu} \quad
  \textbf{Xuyang Cao} \quad
  \textbf{Chao Li} \quad \\
  \textbf{Zhuoyun Liu} \quad
  \textbf{Qintian Sun} \quad
  \textbf{Fangru Zhou} \quad
  \textbf{Haoqiang Xing} \quad
  \textbf{Zhenhong Yang} \quad \\
  \\
  JDH Algo, JD Health International Inc.\\
  \url{https://jdh-algo.github.io/Citrus-V/}\\
}

\begin{document}

\maketitle

\footnoteONLYtext{$^\bigstar$Project Leader.}

\begin{abstract}

Medical imaging provides critical evidence for clinical diagnosis, treatment planning, and surgical decisions, 
yet most existing imaging models are narrowly focused and require multiple specialized networks, limiting their 
generalization. Although large-scale language and multimodal models exhibit strong reasoning and multi-task 
capabilities, real-world clinical applications demand precise visual grounding, multimodal integration, and 
chain-of-thought reasoning. We introduce Citrus-V, a multimodal medical foundation model that combines image 
analysis with textual reasoning. The model integrates detection, segmentation, and multimodal chain-of-thought 
reasoning, enabling pixel-level lesion localization, structured report generation, and physician-like diagnostic 
inference in a single framework. We propose a novel multimodal training approach and release a curated 
open-source data suite covering reasoning, detection, segmentation, and document understanding tasks. 
Evaluations demonstrate that Citrus-V outperforms existing open-source medical models and expert-level imaging 
systems across multiple benchmarks, delivering a unified pipeline from visual grounding to clinical reasoning 
and supporting precise lesion quantification, automated reporting, and reliable second opinions.

\end{abstract}

\section{Introduction}

In clinical practice, physicians routinely operate in highly multimodal 
environments, where medical imaging plays a central role in diagnosis, 
treatment planning, and surgical decision-making. Accurate interpretation of 
imaging data is indispensable, as it provides critical evidence that 
complements textual reports, laboratory results, and patient history. 
Consequently, any artificial intelligence system intended for clinical 
deployment must be capable of integrating visual and textual information at 
a fine-grained, pixel-level resolution while supporting structured reasoning 
and clinically grounded decision-making.

Existing medical imaging models are largely designed as expert systems 
specialized for narrow tasks such as lesion detection\cite{GHESU2018203,10.1016/j.cmpb.2017.06.022,2004Linear}, 
segmentation\cite{zhong2025cross-view,isensee2021nnu,perera2024segformer3d}, 
classification\cite{2004Linear}, or report generation\cite{jing2018on,li2018hybrid,xue2018multimodal,monshi2020survey,unknown2025automated}. 
These models often require multiple specialized networks to cover different 
organs, disease types, or diagnostic tasks, and they rarely generalize 
effectively across diverse clinical scenarios. While large-scale language 
and multimodal models\cite{wang2025internvl3,zhu2025internvl3,team2025kimi,hong2024cogvlm2} 
have demonstrated remarkable progress, including strong reasoning 
capabilities\cite{hong2025glm,su2025thinkingimagesmultimodalreasoning} and 
multi-task generalization, applying them to real-world clinical settings 
remains challenging. Clinical tasks demand not only multimodal understanding 
but also precise visual grounding and integrated chain-of-thought reasoning 
to interpret complex medical data, support decision-making workflows, and 
provide reliable second opinions with explainability and clinical fidelity. 
Existing multimodal medical approaches often fail to provide pixel-level, 
fine-grained visual insights or to integrate heterogeneous data modalities 
effectively, which limits their utility in comprehensive diagnostic 
reasoning.

In our prior work, \textit{Citrus: Leveraging Expert Cognitive Pathways in a Medical Language Model for Advanced Medical Decision Support}\cite{wang2025citrus}, 
we introduced a language-based medical foundation model that incorporated 
expert-inspired reasoning pathways and demonstrated strong performance 
across diverse diagnostic tasks. However, Citrus was fundamentally limited 
to textual modalities and could not directly incorporate medical images, 
which are indispensable for real-world clinical decision-making.

To meet the demands of modern clinical workflows and build upon our previous 
work, next-generation medical foundation models\cite{li2024integrated,liu2025metagp,wong2025eyefm} 
must integrate pixel-level understanding\cite{wu2025unibiomed}, structured 
information extraction\cite{o2015explaining}, and multimodal 
chain-of-thought reasoning\cite{yan2025medreasoner}. By combining robust 
detection and segmentation capabilities, models can achieve precise lesion 
localization and quantitative analyses, such as tumor measurement and tissue 
damage assessment. Integrating these visual insights with textual reports 
enables the generation and interpretation of structured diagnostic 
information, supporting radiology report writing and providing reliable 
second clinical opinions. Furthermore, embedding visual evidence into 
chain-of-thought reasoning allows the model to perform physician-like 
diagnostic inference, improving both accuracy and interpretability.

In this work, we introduce Citrus-V, an upgraded multimodal medical 
foundation model built upon the Citrus framework. Citrus-V makes the 
following key contributions:
\begin{enumerate}
\item We construct a unified model that integrates detection, segmentation, 
and multimodal chain-of-thought reasoning, enabling pixel-level lesion 
localization, structured report generation, and physician-like diagnostic 
inference within a single model.
\item To facilitate reproducibility and support the research community, we 
release Citrus-V along with a curated open-source data suite, including a 
multimodal chain-of-thought reasoning dataset for report generation, a 
refined detection and segmentation benchmark with corrected labels, and a 
medical document understanding benchmark with graded difficulty levels.
\item We further design a novel multimodal training paradigm to accelerate 
convergence and enhance generalization across diverse imaging and reasoning 
tasks.
\end{enumerate}

Extensive experiments demonstrate that Citrus-V surpasses existing 
open-source medical foundation models and expert-level imaging systems 
across multiple benchmarks, establishing new state-of-the-art performance in 
both visual and multimodal tasks. By providing a complete pipeline from 
visual grounding to clinical reasoning, Citrus-V offers critical support for 
precise lesion quantification, automated radiology reporting, and reliable 
second opinions, marking a significant step toward general-purpose medical 
foundation models and the broader adoption of AI in clinical practice.

\section{Related Work}

This section mainly introduces the work related to the contributions of this paper, which will be elaborated from three aspects below, namely medical multimodal foundation models in Section \ref{sec:mmlm}, medical image segmentation foundation models in Section \ref{sec:mism}, and chain-of-thought reasoning technology in Section \ref{sec:ctrt}.

\subsection{Medical Multimodal Large Models}
\label{sec:mmlm}

Traditional medical AI has long relied on expert-designed systems tailored to specific modalities and diseases. For instance, DAMO Academy's research on screening for pancreatic and gastric cancers\cite{cao2023large, hu2025ai} has significantly reduced the cost of detecting major diseases. However, a key limitation is that these approaches require building separate expert models for each imaging modality, disease type, and specific diagnostic task.

Since the introduction of the Transformer architecture\cite{Vaswani2017-pf} and the emergence of ChatGPT\cite{Singhal2025-uu}, large language models(LLM) have been rapidly adapted to the medical domain. 
The Med-PaLM series\cite{Singhal2023-gs, Singhal2025-dj} has distinguished itself by advancing medical question answering (QA) through innovative approaches in data curation, evaluation, and prompting strategies, thereby elevating LLM performance toward physician-level competence. 
Building on open LLaMA architectures, Clinical Camel\cite{Toma2023-vc} introduced conversation-based knowledge encoding, transforming dense medical reviews into synthetic conversational data to facilitate domain-specific supervision. 
Further, IvyGPT\cite{Wang2023-jl} explored staged optimization via supervised fine-tuning and reinforcement learning from human feedback(RLHF), achieving notable performance in Chinese medical QA. 

However, medical data are inherently multimodal. To accommodate the integration of textual and visual information, recent research has increasingly focused on multimodal foundation models. 
Early milestones such as LLaVA-Med\cite{Li2023-hs}, IRENE\cite{2023A}, and Med-PaLM M\cite{Tu2024-ny} demonstrated the feasibility of end-to-end medical vision–language learning. 
Subsequent studies\cite{Hyland2023-ed, Chen2024-lx, Zhang2024-yu, Li2024-an, LASA_Team2025-yp, Sellergren2025-ly} have scaled both datasets sizes and model capacities, collectively marking a transition from task-specific solutions toward general-purpose medical multimodal large models (MMLMs).

Concurrently with the steady progress in MMLMs, the supported visual modalities have expanded from initially only 2D images to include 3D imaging data.
Pioneering efforts like MedBLIP\cite{Chen2024-nj} enabled efficient fusion between 3D medical images and text, while RadFM\cite{Wu2025-rp} introduced a unified generative architecture supporting multiple tasks, addressing the scarcity of general-purpose foundational models in radiology. 
More recent work, including M3D\cite{Bai2024-vg}, Dia-LLaMA\cite{Chen2024-mm}, and Med-2E3\cite{Shi2024-un}, has further advanced data strategies, task formulations, and model architectures, accelerating the translation of 3D MMLMs into clinical applications.

Despite this progress, two key challenges remain: the scarcity of paired 3D image–report data, and the quadratic complexity of modeling long 3D sequences.

\subsection{Medical Image Segmentation Models}
\label{sec:mism}
Segmentation has been a cornerstone of medical imaging. CNN-based architectures like U-Net\cite{ronneberger2015u} and its successors\cite{cao2020uncertainty, zhou2019unet++} capture multi-scale contextual information 
through skip connections, in addition, nnU-Net\cite{isensee2021nnu}, which further demonstrates the power of self-adaptive pipelines, has set benchmarks across diverse datasets. Transformer-based extensions such as UNETR\cite{hatamizadeh2022unetr} and MedNeXt\cite{roy2023mednext}, improving long-range modeling, offer robustness in complex clinical scenarios but remain limited by annotation requirements and adaptability.

The introduction of foundation segmentation models such as SAM\cite{kirillov2023segment} and SAM2\cite{ravi2024sam} has motivated medical adaptations like 
MedSAM\cite{ma2024segment} and MedSAM2\cite{zhu2024medical, ma2025medsam2}. By leveraging visual or textual prompts, these models provide interactive solutions 
for delineating medical structures, often requiring minimal user guidance. While highly effective for semi-automatic workflows, such methods depend heavily on prompt 
quality and lack robustness in large-scale fully automated settings. This motivates research into more tightly coupled multimodal frameworks that can integrate domain 
priors with general foundation models.

Integrating large language models (LLMs) with vision encoders has opened new paradigms for segmentation. One direction formulates segmentation as text generation, 
where models such as Florence-2\cite{xiao2024florence} and VisionLLM\cite{wang2023visionllm} represent masks through language tokens, though achieving pixel precision 
remains difficult. Another line introduces task-specific decoders, pioneered by LISA\cite{lai2024lisa}, which employs a <SEG> token aligned with LLM hidden states. 
Extensions including LISA++\cite{yang2023lisa++}, GLAMM\cite{rasheed2024glamm}, SA2VA\cite{yuan2025sa2va}, and X-SAM\cite{wang2025x} expand this design to multi-modal 
and zero-shot segmentation. Finally, unified frameworks such as UFO\cite{tang2025ufo} and Pixel-SAIL\cite{zhang2025pixel} aim to integrate detection, segmentation, and 
pixel understanding within a single transformer architecture, pointing toward scalable and general-purpose solutions.

Extending these ideas to the clinical domain, recent works combine LLMs with medical encoders for multimodal analysis. M3D\cite{bai2024m3d} targets 3D medical segmentation, 
while MIMO\cite{chen2025mimo} and MedPliB\cite{huang2025towards} enhance interpretability by coupling segmentation with region-level VQA. UniBioMed\cite{wu2025unibiomed} 
further unifies segmentation and diagnostic report generation, offering joint localization and reasoning capabilities. Most approaches adopt parameter-efficient 
fine-tuning (e.g., LoRA\cite{hu2022lora}) to preserve language capacity, but this can limit multi-task synergy. By contrast, our method employs full-parameter supervised fine-tuning, 
encouraging deeper integration of pixel-level segmentation and medical language understanding within a single multimodal large model.

\subsection{Chain-of-Thought Reasoning Technology}
\label{sec:ctrt}

Large language models have increasingly been found to be prone to hallucination, generating plausible yet nonfactual content. Recently, in-context learning (ICL) techniques empower large language models to demonstrate stepwise reasoning, typically referred to as Chain-of-Thought (CoT) reasoning mechanisms. 
This approach enables models to break down problems into a series of intermediate steps, where a considerable amount of previous research\cite{wei2022chain,kojima2022large,zhang2022automatic,zhou2022least,lu2022dynamic,guo2025deepseek} has concentrated on, improving performance on complex reasoning tasks while further enhancing transparency in the model's decision-making process. 

However, existing studies\cite{wei2022chain,wang2022self,zhou2022least,chen2022program,ho2022large} related to CoT reasoning are largely isolated in the language modality, which has been widely applied to tasks such as mathematical reasoning, commonsense reasoning, and symbolic reasoning. Studies have shown that CoT can not only improve the model's accuracy on standard benchmark tests (e.g., GSM8K, AQUA-RAT) but also enhance the transparency and interpretability of the model's reasoning process.

Inspired by the success of models like DeepSeek~R1\cite{guo2025deepseek} and OpenAI o3\cite{openai2025o3}, researchers are actively exploring how to apply similar reasoning enhancement methods to multimodal large language models (MLLMs). By combining textual and visual information to train models for step-by-step reasoning, Multimodal Chain-of-Thought (MCoT) enables models to better focus on critical regions within visual inputs and establish connections between text and images. 
Representative examples of previous work include, Multimodal-CoT\cite{zhang2023multimodal}, MVoT\cite{li2025imagine}, Video-of-Thought\cite{fei2024video}, Audio-CoT\cite{ma2025audio}, PCoT\cite{wang2024t}, LLaVA-CoT\cite{xu2024llava}, CCoT\cite{mitra2024compositional}, UV-CoT\cite{zhao2025unsupervised}, VisCoT\cite{shao2024visual}. 
To some extent, MCoT mitigates hallucinations and facilitates model convergence.

Unlike general-purpose multimodal models, multimodal models in the medical domain place greater emphasis on the interpretability of the CoT reasoning process.
To endow models with physician-level or even superior performance in clinical diagnosis, treatment planning, and surgical decision-making, the models must possess sophisticated cognitive reasoning capabilities, specifically the ability to perform iterative reasoning through contextual understanding and self-correction. Models using MCoT can accurately replicate the doctor's diagnosis and treatment process\cite{yan2025medreasoner, kyung2025medregion}. They generate diagnostic reports by integrating the image features of lesion areas and medical history texts, which greatly improves the accuracy and efficiency of image diagnosis.

\section{Data Curation}

In this section, we collecte and curate datasets sourced from open-source repositories to ensure diversity in instructions and comprehensiveness in the medical image-caption corpus. Additionally, we include synthesized data specifically designed to improve medical reasoning capabilities and enhance the model’s performance across various downstream task domains.
\subsection{Data Collection}

\subsubsection{Data Source}

To enhance the fine-grained visual understanding capabilities of Citrus-V in medical imaging and facilitate curriculum learning for medical instruction tuning, we curated and constructed a curriculum dataset from open-source resources. This curriculum comprises medical image captioning data, medical instruction tuning data, and multimodal data from both medical and general domains, including interpretation tasks involving text, charts, diagrams, and scientific illustrations extracted from journal articles.

\paragraph{Medical Image-caption Data}The open-source datasets like PMCOA\cite{lin2023pmcclipcontrastivelanguageimagepretraining}, ROCO\cite{ROCO}, contain a large collections of coarse caption of medical images, contributive to align with diversity of medical conceptions, covering different modalities and diseases in biomedical category. Medical image reports are considered as a more comprehensive image-captioning corpus, rigorousely curated and vetted by experts, maintained by hospital or health institutions, written by professional radiologists, or taken from informative medical science documentaries. Systematic radiology report formulation, detailed perception and interpretation of medical images are extensively summarized and processed for intergration. The following dataset were assembled: CheXpert Plus\cite{chambon2024chexpertplusaugmentinglarge}, PubMedVision\cite{chen2024huatuogptvision}, LLaVA-Med\cite{li2023llava}, ROCOv2\cite{ROCOv2}, IU-Xray\cite{demner2016preparing-iu-xray}, MIMIC-CXR\cite{johnson2019mimiccxr}. 

\paragraph{Medical MM Instruction Data}Medical image interpreting tasks can be categorized into several distinct types, including anatomical organ identification, imaging features interpretation, lesion localization, diagnosis and symptom inference. The diverse Medical Instruction data comes from open-source dataset including VQA-RAD\cite{lau2018dataset}, PathVQA\cite{naseem2022vision}, PMC-VQA\cite{zhang2023pmc}, SLAKE\cite{liu2021slake}, MIMIC-Ext-MIMIC-CXR-VQA\cite{bae2024ehrxqa}, VQA-Med-2019\cite{ImageCLEFVQA-Med2019}, supplemented to enrich specific task categories, are curated to enhance the systematicity, comprehensiveness and the reasoning ablility while completing similar task.

\begin{table}[t]
  \centering
  \caption{List of Open-Source Dataset Collected. We aggregate public corpora across four categories—medical image–caption, 
  medical multimodal instruction/VQA, medical textual QA and reasoning, and general instruction—to ensure diverse instructions 
  and comprehensive coverage for medical vision–language training. Representative sources include PMCOA/ROCO, LLaVA-Med, MIMIC-CXR(+VQA), 
  VQA-RAD/PMC-VQA, CheXpert Plus, MedQuAD, JMed, and OpenHermes; the full list appears in the table.}
  \label{tab:dataset-included}
  \resizebox{0.95\textwidth}{!}{%
  \normalsize
  \begin{tabular}{l p{10cm}}
  \toprule
  \textbf{Aspect} & \textbf{\# Datasets Included} \\
  \midrule
  
  \colorbox{yellow!12}{Medical Image-Caption} & PMCOA, ROCO, LLaVA-Med, MedPix2.0, CheXpert Plus, MIMIC-CXR, ROCOv2, Quilt-LLaVA, PubMedVision\\
  \arrayrulecolor{gray!50}
  \cmidrule(lr{1em}){1-2}
  \arrayrulecolor{black}
  \colorbox{yellow!12}{Medical MM Instruction}   & VQA-RAD, PMC-VQA, PATH-VQA, SLAKE, MIMIC-CXR-VQA, VQA-Med-2019\\
  \midrule
  \multirow{5}{*}{\colorbox{yellow!12}{Medical Textual Data}} & JMED, HealthCareMagic, icliniq, HuatuoGPT2-SFT-GPT4\\
  \arrayrulecolor{gray!50}
  \cmidrule(lr{1em}){2-2}
  \arrayrulecolor{black}
   & Citrus-S3, medical-o1-verifiable-problem, Medical-R1-Distill-Data, huatuogpt-o1-for-reasoning, MedReason, MedThoughts, medical-o1-reasoningSFT\\
   \arrayrulecolor{gray!50}
   \cmidrule(lr{1em}){2-2}
   \arrayrulecolor{black}
   & AlpaCare, ApolloCorpus, MedQuAD, MedQA, PMC-LLaMA\\
  \midrule
  \colorbox{yellow!12}{General Instruction} & LLaVA1.5, PixMo, ALLaVA, OpenHermes-2.5 \\
  \bottomrule
  \end{tabular}
  }
\end{table}
\paragraph{Medical Textual Data}A substantial amount of medical textual data originates from professional medical examination questions as normal medical instructions including AlpaCare\cite{zhang2025alpacareinstructiontunedlargelanguagemodels}, ApolloCorpus\cite{wang2024apollo}, MedQuAD\cite{BenAbacha-BMC-2019},\cite{jin2021disease}, PMC-LLaMA\cite{wu2023pmcllamabuildingopensourcelanguage}, anonymized patient-doctor dialogue and medical consultations, including HealthCareMagic\cite{li2023chatdoctormedicalchatmodel},icliniq-10k\cite{li2023chatdoctormedicalchatmodel}, HuatuoGPT2-SFT-GPT4\cite{chen2023huatuogptii}, and also records released in our previous work JMED\cite{wang2025citrusleveragingexpertcognitive}, and medical textual reasoning data distilled, including Citrus-S3\cite{wang2025citrusleveragingexpertcognitive}, medical-o1-verifiable-problem\cite{chen2024huatuogpto1medicalcomplexreasoning}, Medical-R1-Distill-Data\cite{chen2024huatuogpto1medicalcomplexreasoning}, huatuogpt-o1-for-reasoning\cite{chen2024huatuogpto1medicalcomplexreasoning}, MedReason\cite{wu2025medreasonelicitingfactualmedical}, MedThoughts-8K\cite{medthoughts8k}, medical-o1-reasoningSFT\cite{chen2024huatuogpto1medicalcomplexreasoning}, was incorporated and filtered to mitigate potential biases arising from image-dominated multimodal training and enhances the model’s performance on downstream tasks that rely primarily on textual understanding and reasoning.
\paragraph{General Instruction Data}In addition, textual and multimodal in general fields are introduced to further improve the performance in locating fine-grained visual details, open-knowledge VQA tasks and additional academic task-oriented datasets. The open-source datasets taken from OKVQA\cite{marino2019okvqavisualquestionanswering}, A-OKVQA\cite{schwenk2022aokvqabenchmarkvisualquestion}, OCRVQA\cite{mishraICDAR19}, TextCaps\cite{sidorov2019textcaps}, involved and have been improved in the previous work of LLaVA1.5\cite{liu2024improvedbaselinesvisualinstruction}. PixMo\cite{deitke2024molmopixmoopenweights} contains a diverse array of tasks involving the interpretation and reasoning over diagrams, charts, and illustrations, thereby supporting comprehensive understanding of the charts and tables in the document. OpenHermes2.5\cite{OpenHermes2-5} and ALLaVa\cite{chen2024allava} contains high-resolution images with fine-grained captions and complex instructions, utilized to supplement the training corpus with more diverse and complex long-form instructions, thereby enhancing the model’s capabilities.

\subsubsection{Data Process}
\begin{figure}[t!]
    \centering
      \includegraphics[width=1.0\textwidth]{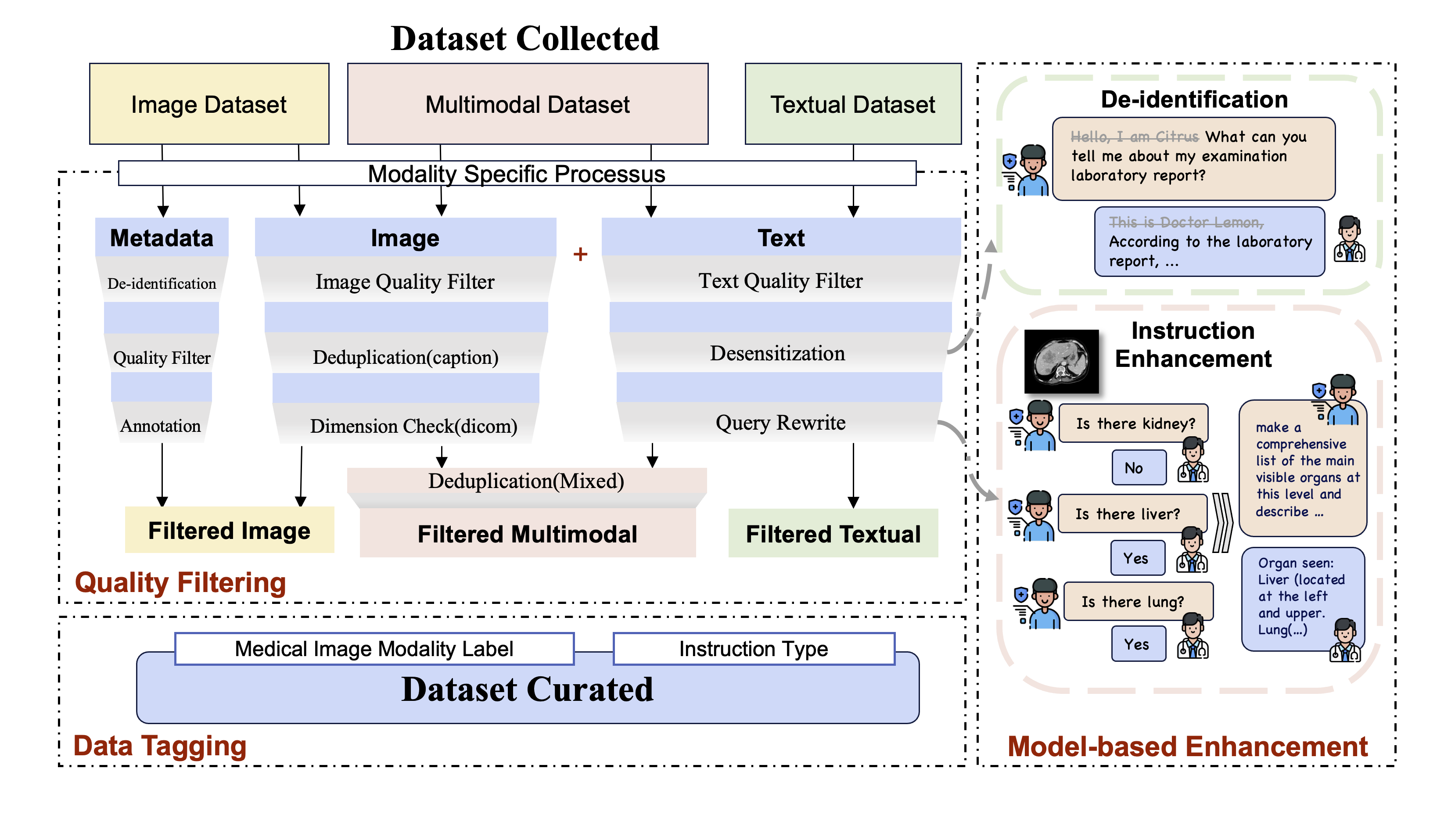}
    \caption{Overview of the Open-source Dataset Filtering and Processing Pipeline. The image and textual data quality is assesse, following with filtering and task-level deduplication, the model-based method are incorporated into the processing pipeline to perform de-identification and instruction enhancement; }
    \label{fig:dataset_prep}
  \end{figure}

To ensure the quality and utility of the collected comprehensive dataset, we implemented a data processing pipeline, which involves basic filtering methods to remove low-quality or irrelevant samples, and employs model-based enhancement techniques to further refine and enrich the dataset. The following sections details the procedures and criteria adopted. The complete data curation pipeline is illustrated in the Figure~\ref{fig:dataset_prep}, which details the procedures and criteria adopted.
\paragraph{Quality Filtering}
After extensively collecting the aforementioned open-source datasets, rigorous data processing pipeline was applied to both textual(unimodal), image(unimodal) and image-text (multimodal) data. Initially, the quality of images, the metadata included and texts was carefully assessed based on several criteria.

For \textbf{image data} in both medical and general domains, 4096 pixels is used as a criterion to filter out low-quality images. For expert-level medical imaging data designated for further synthesis, which are stored in DICOM format with \textbf{metadata}, additional criteria—including physical dimension consistency, orientation consistency, and slice-spacing consistency—are applied to ensure data quality. Files exhibiting errors in orientation metadata, specifically in the anterior-posterior, left-right, or superior-inferior directions, are excluded. Furthermore, all personal or private information contained in the metadata is thoroughly removed as part of the data de-identification process.

As for \textbf{textual data} sample, low-quality samples were primarily filtered out using methods including regex and token thresholds to remove the sample too short, fewer than 10 or more than 1024 tokens, preventing from the potential no-relevance data, especially for the texutal part which concentrates on the image captioning and report generation tasks. Deduplication was also applied to pure text instruction–response pairs to prevent excessive redundancy within the integrated dataset. 

In the case of \textbf{multimodal} samples, in addition to image and text quality filtering and deduplication, duplication of image-text pairs was identified and removed by combining textual and image hash results, thereby ensuring the uniqueness and overall quality of the dataset.

\paragraph{Model-based Enhancement}

Model-based methods were applied to portions of the filtered dataset, primarily targeting samples with relatively low instruction and response quality, spotting in the trainset contained in the open-source VQA dataset and for those containing personal privacy information independent from the topic, such as data curated from doctor-patient dialogues or health consultations.

\textbf{De-identification:} Conversational data, originating from real user inputs, can significantly enhance the diversity of the training dataset. However, personal and sensitive information present in these dialogues is often not directly relevant to the underlying medical issues. To ensure the quality and privacy compliance of the training data, for the textual datasets derived from conversational data, to perform de-identification, our strategy involve leveraging large language models (LLMs) to anonymize and rewrite the content, thereby removing personal information unrelated to medical consultation from doctor-patient exchanges, we employed LLMs to systematically remove such information, retaining only content pertinent to medical consultation.

\textbf{Instruction Enhancement:} VQA-type training and evaluation datasets are commonly organized into several categories: short-answer questions, multiple-choice questions, true/false questions, and open-ended questions (distinguished from short-answer questions). 

The first issue arises from the limited availability of public medical instruction tuning or VQA data, the scarcity of accompanying labels, and the rarity of expert annotations. As a result, some training sets are constructed by posing multiple questions about the same image, or by including a large number of binary presence/absence queries (e.g., “Is the organ present?”) and various organ identification and localization tasks. 

The second issue concerns the nature of responses in most training data: although answers in the training set samples are all provided with minimal word count, the instructions do not explicitly require concise responses. Furthermore, for a given question, homogeneous, similar, or even different perspectives can yield equally valid answers, indicating that while these responses are succinct and correct, they may lack systematicity and comprehensiveness. 

For such instruction types, the majority of visual understanding questions may be overly simplistic for models that already possess medical visual capabilities, offering limited benefit for training on complex tasks such as reasoning and visual relationship inference. Therefore, we consider the primary utility of this data to be in improving the model’s ability to follow specific instructions rather than enhancing its reasoning or advanced visual understanding capacity.

Thus, our main object was to enable the model to provide more systematic and comprehensive responses to image-related queries by instruction enhancement. To this end, we performed further processing on datasets featuring brief and simple responses by utilizing a VLM to consolidate multiple questions about a single image into more holistic and focused queries, thereby enhancing the professionalism and depth of image interpretation and relational reasoning across multiple image features in medical imaging in the model’s outputs.

\textbf{Annotations:} For the imaging data, organ and lesion-level annotations are also generated using registration-based methods, which assign labels ang ROIs to specified regions within the filtered imaging data. The detailed workflow for this process is described in the following section.

\paragraph{Data Tagging}

The processed data encompasses a wide range of instruction types, each targeting diverse task categories and imaging modalities within medical images. Data tagging methods were extensively employed to annotate both training and evaluation datasets with corresponding task types. The diversity of instruction formats and the broad coverage of imaging modalities and task categories in the training data contribute to enhancing the model’s performance across a variety of downstream tasks. 

To analyze the distribution of different data and clearly illustrate the data distribution and better assess the model’s performance across diverse categories, we systematically annotate all medical data according to imaging modality, task type, and anatomical region. This comprehensive tagging scheme not only provides a transparent overview of the dataset’s composition, but also facilitates a detailed evaluation of the model’s capabilities for each task tag, while highlighting potential biases present within the data. These insights inform targeted refinements of both the dataset and the model. The specific classification scheme is shown in Table \ref{tab:label_class}.

\begin{table}[t]
\centering
\caption{All Labels Employed in the Datasets. Data are tagged by imaging modality, 
task, and anatomical region to enable stratified analysis. 
The table lists all labels used for each dimension.}
\label{tab:label_class}
\begin{tabular}{l|m{0.7\textwidth}}
    \toprule
    \textbf{Dimension} &\textbf{Data Tag}  \\
    \midrule
     Modality   &  Histopathology, CT, X-Ray, MRI, NIR, Ultrasound, Microscopy, OCT, Dermoscopy, Photograph, Endoscopy, Fundus, Other\\
     \midrule
     Task   &  Examination Selection, Report Analysis, Prior Comparison, Modality Recognition, Organ Recognition, Image Description, Report Generation, Lesion Localization, Differential Diagnosis, Symptoms Inference, Treatment Generation, Treatment Selection, Treatment Details, Basic, Other\\
     \midrule
     Anatomical Region   & Abdomen, Chest, Brain, Neck, Cell, Lower Limb, Upper Limb, Oral Cavity, Eye, Breast, Gastrointestinal Tract, Pelvis, Foot, Joint, Other \\
    \bottomrule
\end{tabular}
\end{table}

When tagging imaging modalities, a dedicated model is trained to classify images by modality. For task and anatomical region, an LLM is utilized to assign tags. In particular, task tagging requires additional domain expertise. Therefore, a detailed strategy is implemented in which precise definitions and representative examples for each task tag are provided to the LLM to ensure accurate annotation.  This curated set of reasoning annotations was then utilized to further enhance the model’s capacity for systematic and robust reasoning as shown in Figure \ref{fig:vqa-reasoning}.

\begin{figure}[t!]
    \centering
      \includegraphics[width=1.0\textwidth]{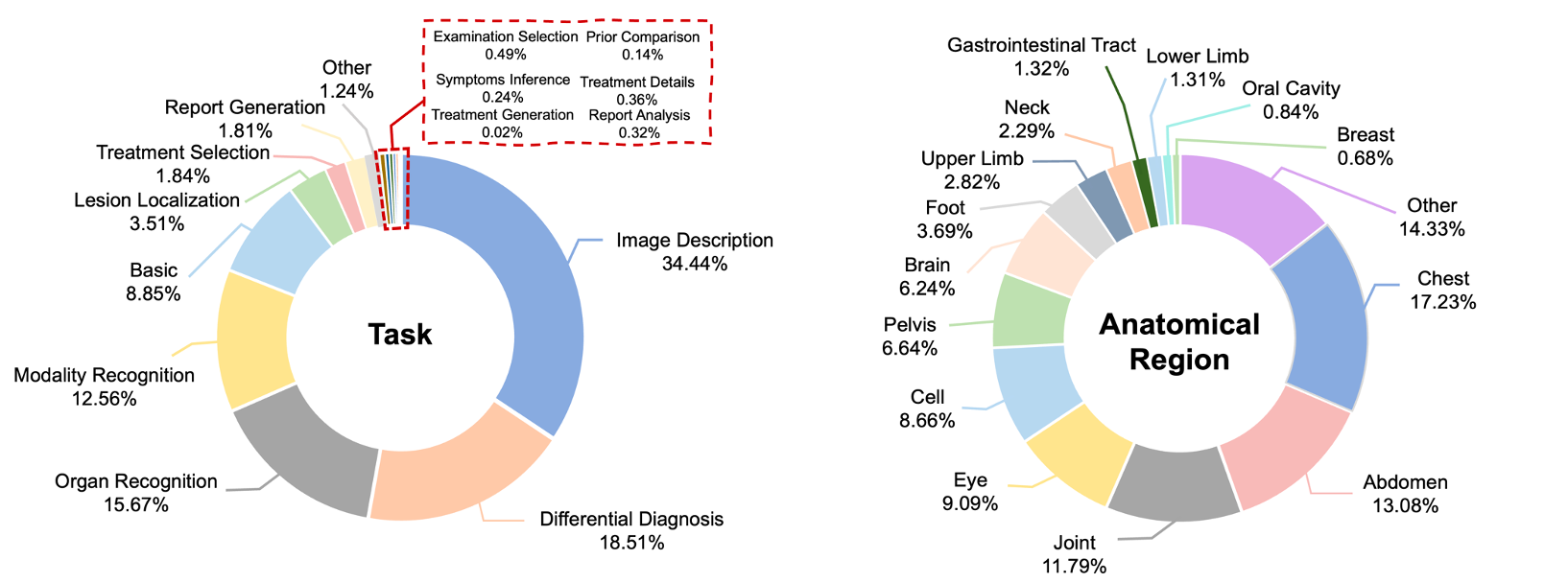}
    \caption{Distribution of Tagged Data. The task tag summarizes the distribution of both text and multimodal test data, while the anatomical region tag summarizes the distribution of multimodal data only.}

    \label{fig:test-distribution}
  \end{figure}

The modality distribution of data is shown in Section \ref{sec:TrainStage}, while the task and anatomical region distributions are shown in Figure \ref{fig:test-distribution}. We summarize the distribution of all data in task. Since pure text data do not have corresponding medical images, the statistics for modality and anatomical region are limited to multimodal data only.

\subsection{Data Synthesis}
\label{sec:datasynthesis_all}
\subsubsection{Medical Visual Reasoning}
\begin{figure}[t!]
    \centering
      \includegraphics[width=1.0\textwidth]{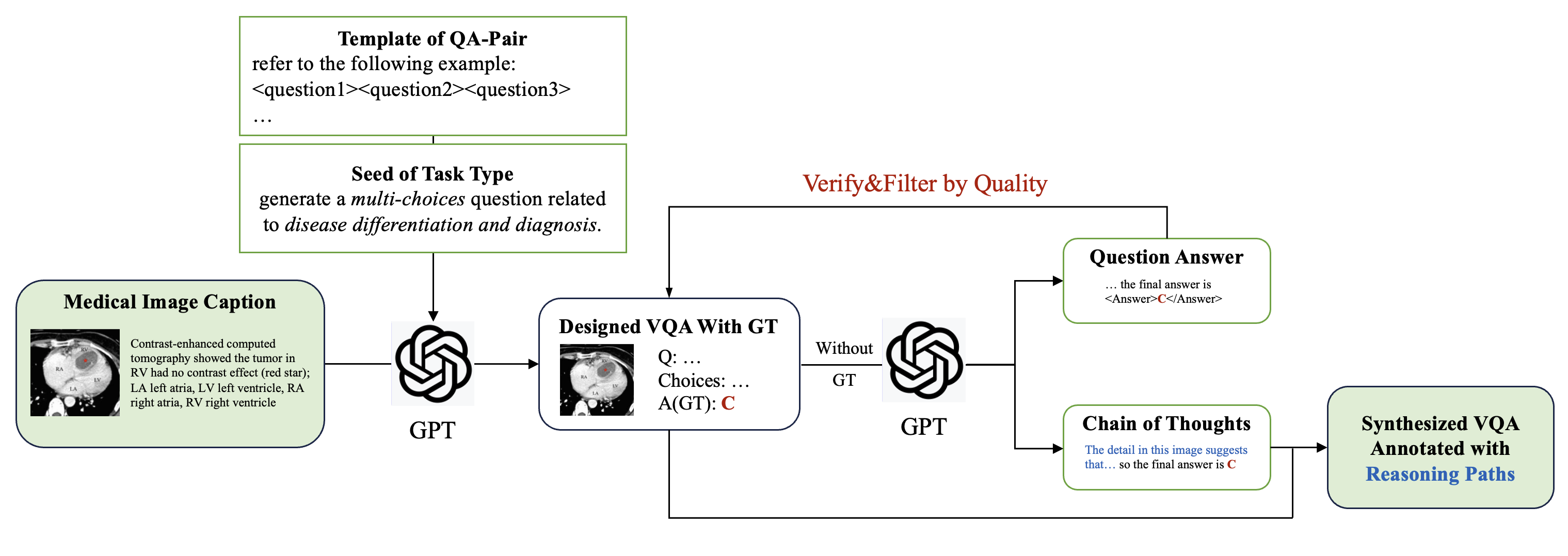}
    \caption{Data Pipeline for Medical Visual Reasoning, different LLMs were employed to (1) generate question-answer pairs with standard answers, tailored according to specific task types, question formats, and image information. (2) annotate the reasoning processes underlying each question-answer pair, thereby verifying the existence of clear and logical reasoning paths, only those reasoning processes that aligned with the designed question answers were retained, ensuring the quality and validity of the annotated data. }

    \label{fig:vqa-reasoning}
  \end{figure}

  In existing multimodal question-answering datasets, the QA format often features relatively short and concise responses, typically limited to multiple-choice, true/false, or short-answer questions, and generally lacking in systematic and comprehensive coverage. While concise question-answer pairs are commonly employed for perception and grounding tasks, the training data also includes QA formats that require more complex reasoning. For such tasks, the absence of explicit reasoning paths may hinder the model’s performance on multimodal tasks that demand intricate inference across heterogeneous information sources.

  To address this, we filtered medical VQA queries by response length and query difficulty, and utilized the pipeline illustrated in Figure \ref{fig:vqa-reasoning} to synthesize medical VQA reasoning data. This process focuses on specific task types and incorporates challenging VQA items from the training set, ensuring that these complex problems are accompanied by clear analytical pathways leading to correct answers through reasoning.
  
  We employed both a generation model and an answering model to respectively synthesize VQA and VQA reasoning data. Following data annotation, model performance across different data tags was analyzed to highlight specific categories. Representative examples were sampled from the annotated data, and specialized prompts were designed to synthesize the required VQA data. Throughout dataset construction, verification procedures ensured that the predefined reasoning paths consistently yielded accurate solutions to complex problems.

\subsubsection{Medical Document Understanding}
\label{sec:MedDocTrain}

\begin{figure}[t!]
  \centering
    \includegraphics[width=\linewidth,trim=2.3cm 0 4.4cm 0,clip]{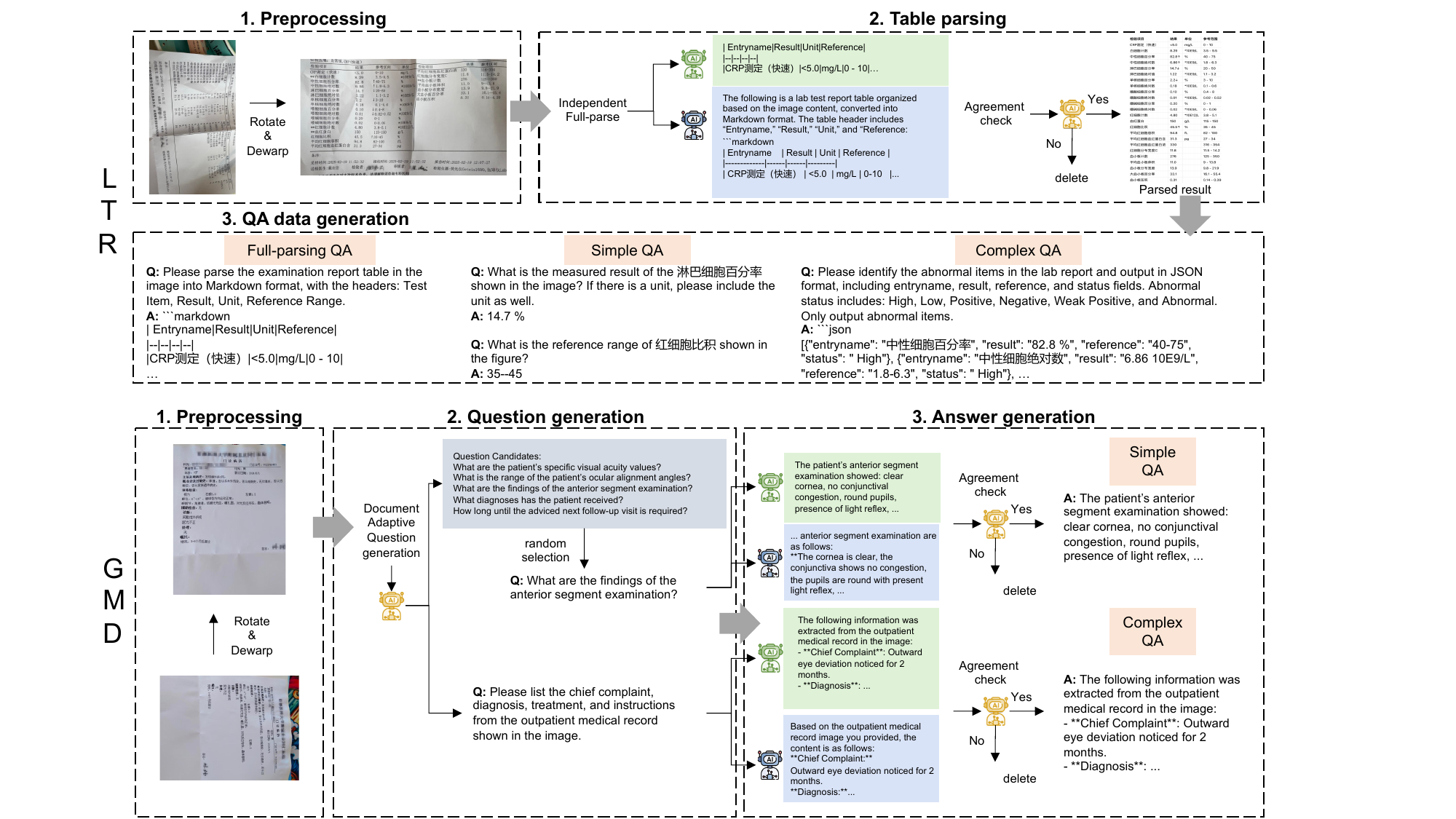}
    \caption{Data Pipeline for Medical Document Understanding. The figure depicts two pipelines—one for Laboratory Test Reports 
    (LTR) and one for General Medical Documents (GMD). Both pipelines apply page-geometry normalization, 
    dual parsing, and LLM-based agreement checking to construct high-quality supervision and mine “hard” cases that require 
    normalization to reach consensus.}
  \label{fig:meddoc_pipeline}
\end{figure}

During online or offline consultations, laboratory results, imaging reports, prescriptions, and clinical notes are 
frequently presented. Public datasets, however, rarely contain real-world text-rich medical document images. 
To bridge this gap, we curate a corpus of \(\sim\)2M images uploaded by Chinese patients in routine online 
consultations and build an automated labeling pipeline (Figure~\ref{fig:meddoc_pipeline}) using state-of-the-art OCR 
and vision–language models (VLMs) to extract text, producing QA datasets as training data and a held-out benchmark. 
Data distribution is in Table~\ref{tab:meddoc-data-dist}.

\begin{figure}[t!]
  \centering
    \includegraphics[width=1.0\textwidth]{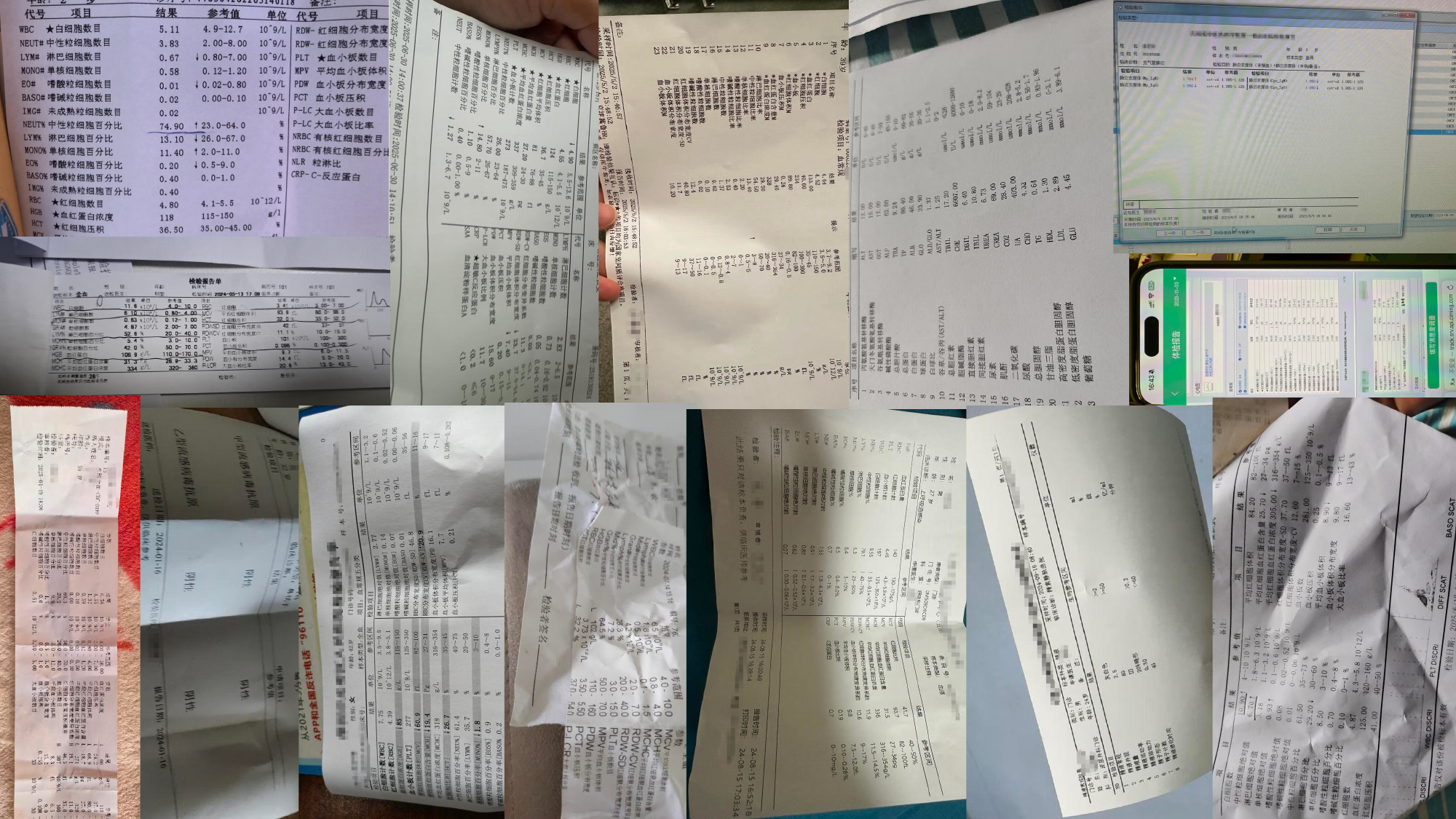}
  \caption{Laboratory Test Report Example Images. 
  Representative cases illustrate heterogeneous templates, dense tabular layouts, and common capture artifacts 
  (skew, blur, partial crops, shadows, folds, and low contrast). These factors hinder reliable cell detection 
  and value alignment and complicate abnormality identification, which requires precise numeric comparison 
  against reference intervals.}
  \label{fig:labtest}
\end{figure}

\paragraph{Laboratory Test Reports (LTR)} 
Current VLMs struggle most on laboratory test reports (LTR), which represent a large portion of our corpus (second only to 
medication-package images) and are clinically critical. The difficulty arises from (i) heterogeneous layouts and dense 
tabular content; (ii) image quality issues (partial crops, blur, skew/rotation, warping, folds); and (iii) the need 
for precise numeric comparison against reference intervals to identify abnormalities, which is harder than generic 
table parsing. Figure~\ref{fig:labtest} shows typical artifacts. 

Our pipeline begins with preprocessing to detect orientation and geometric distortion, applying rotation and dewarping 
to normalize each page. We then perform table parsing with two state-of-the-art VLM parsers that convert each report 
into a four-column Markdown table \texttt{(entry\_name, result, reference, unit)}; after rule-based normalization, 
an LLM agreement checker retains only exact or semantic matches. From the normalized tables we derive three types of 
question-answering (QA) pairs: (i) full parsing, (ii) simple QA that queries a single field, and (iii) complex/abnormality QA 
that compares \texttt{result} against \texttt{reference}. For hard-example mining, images are labeled as Easy 
(Consensus-Before)—consistent full parses obtained without rotation/dewarping—or Hard (Consensus-After)—initially 
inconsistent parses that become consistent after normalization. Models trained with a higher proportion of Hard samples
generalize better to in-the-wild inputs, so our post-training set prioritizes these examples.

\begin{figure}[t!]
  \centering
    \includegraphics[width=1.0\textwidth]{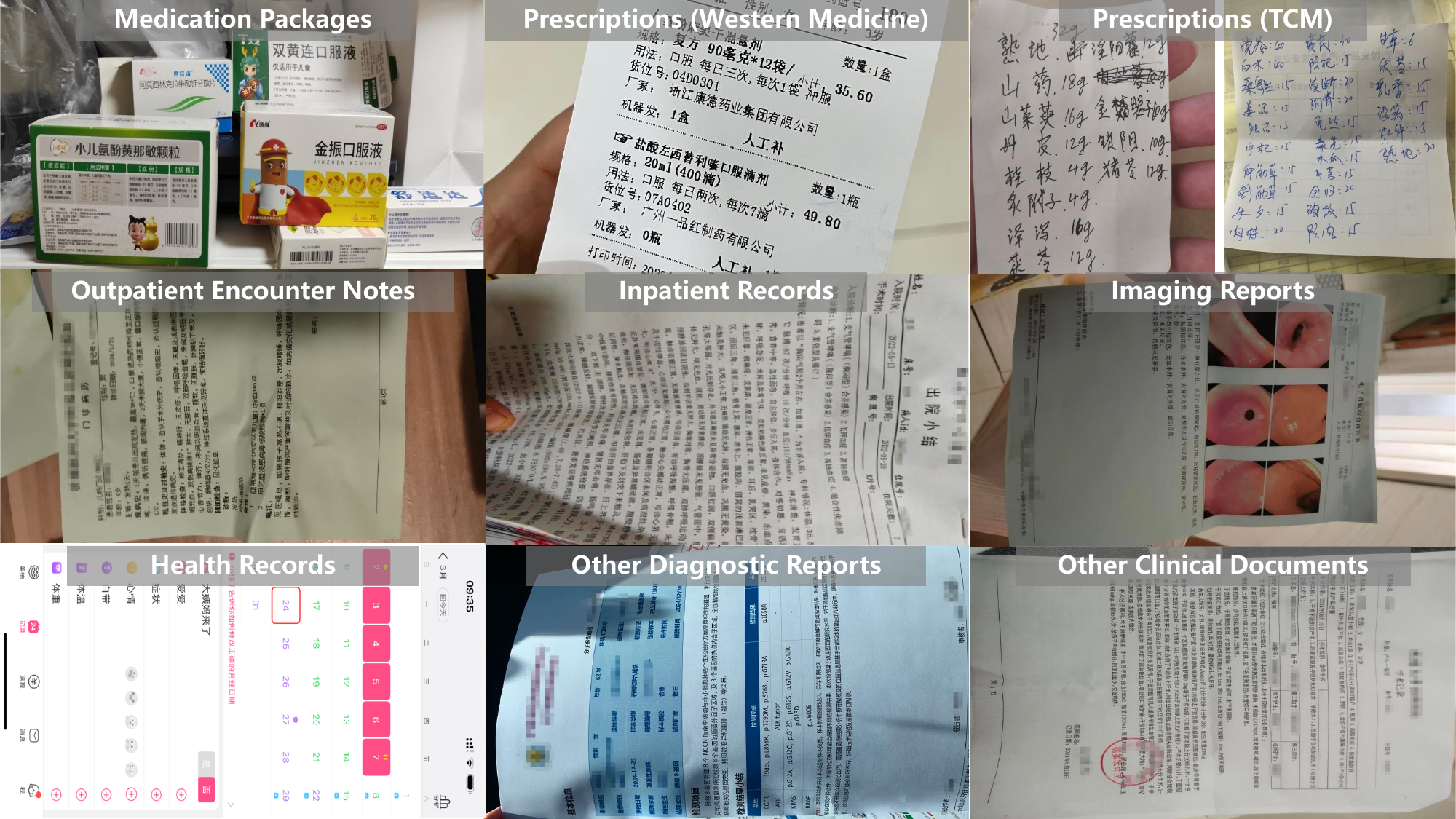}
  \caption{General Medical Document Example Images. Documents exhibit diverse formats, typography, and sectioning conventions. 
  This variability motivates a document-type–adaptive QA formulation rather than a single full-parse schema, while retaining the 
  same geometry normalization, dual extraction, and agreement filtering stages.}
  \label{fig:meddoc}
\end{figure}

\paragraph{General Medical Documents (GMD)}
Beyond LTR, we extend the pipeline to general medical documents (imaging reports, prescriptions, discharge 
summaries, etc.). We retain the same core steps—geometry normalization, dual-parser extraction, and LLM agreement 
checking—but refrain from enforcing a single full-parse schema due to greater format heterogeneity and diverse clinical 
intents (see Figure~\ref{fig:meddoc}). Instead, we construct document-type–adaptive QA supervision.

\begin{figure}[t!]
    \centering
    \begin{subfigure}[b]{0.49\textwidth}
        \centering
        \includegraphics[width=\textwidth,trim=0.15cm 0 0.15cm 1,clip]{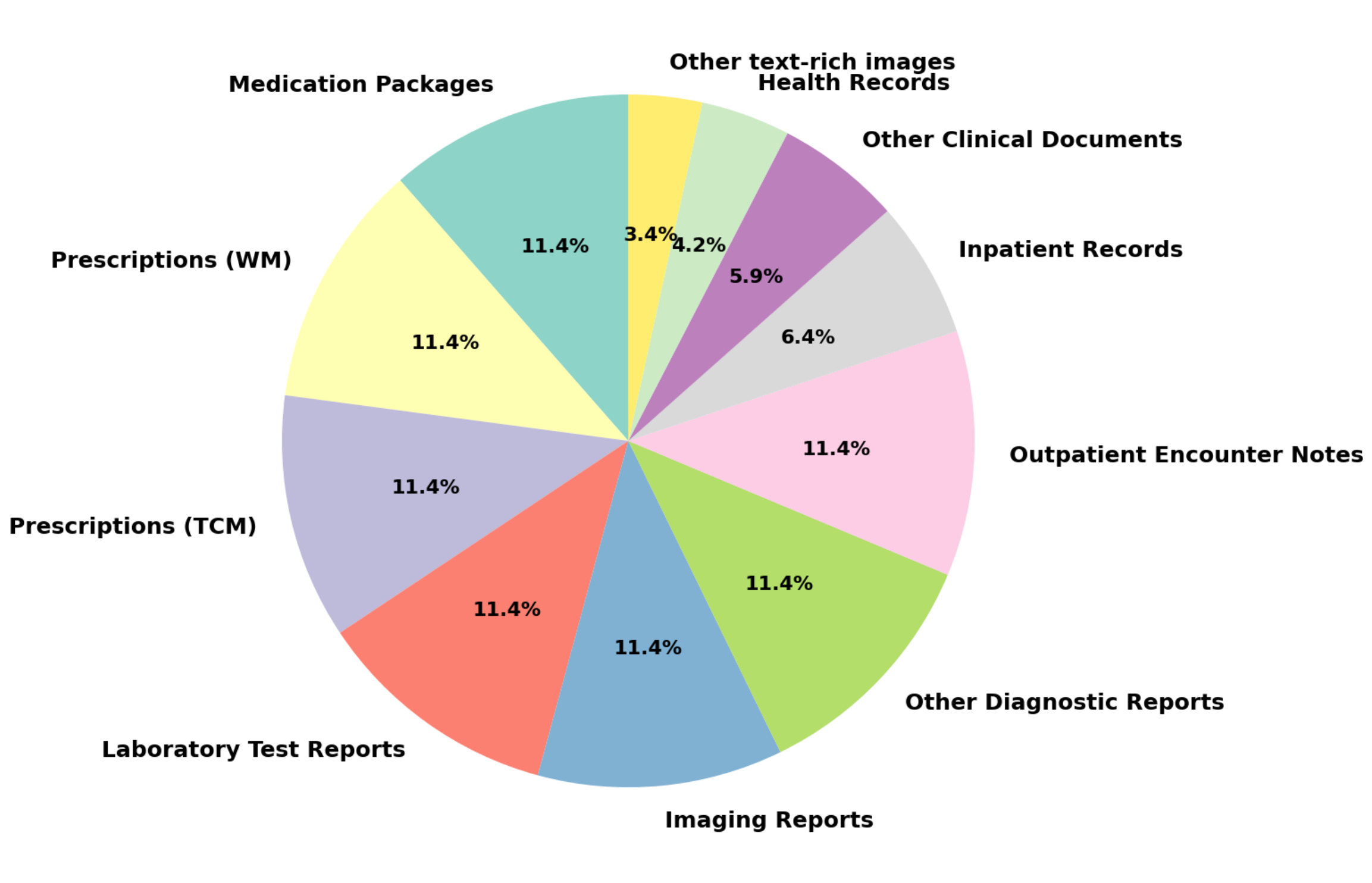}
        \caption{Distribution of GMD Training Data.}
        \label{fig:meddoc_dist}
    \end{subfigure}
    \hfill
    \begin{subfigure}[b]{0.45\textwidth}
        \centering
        \includegraphics[width=\textwidth,trim=0.2cm 1cm 0.2cm 0,clip]{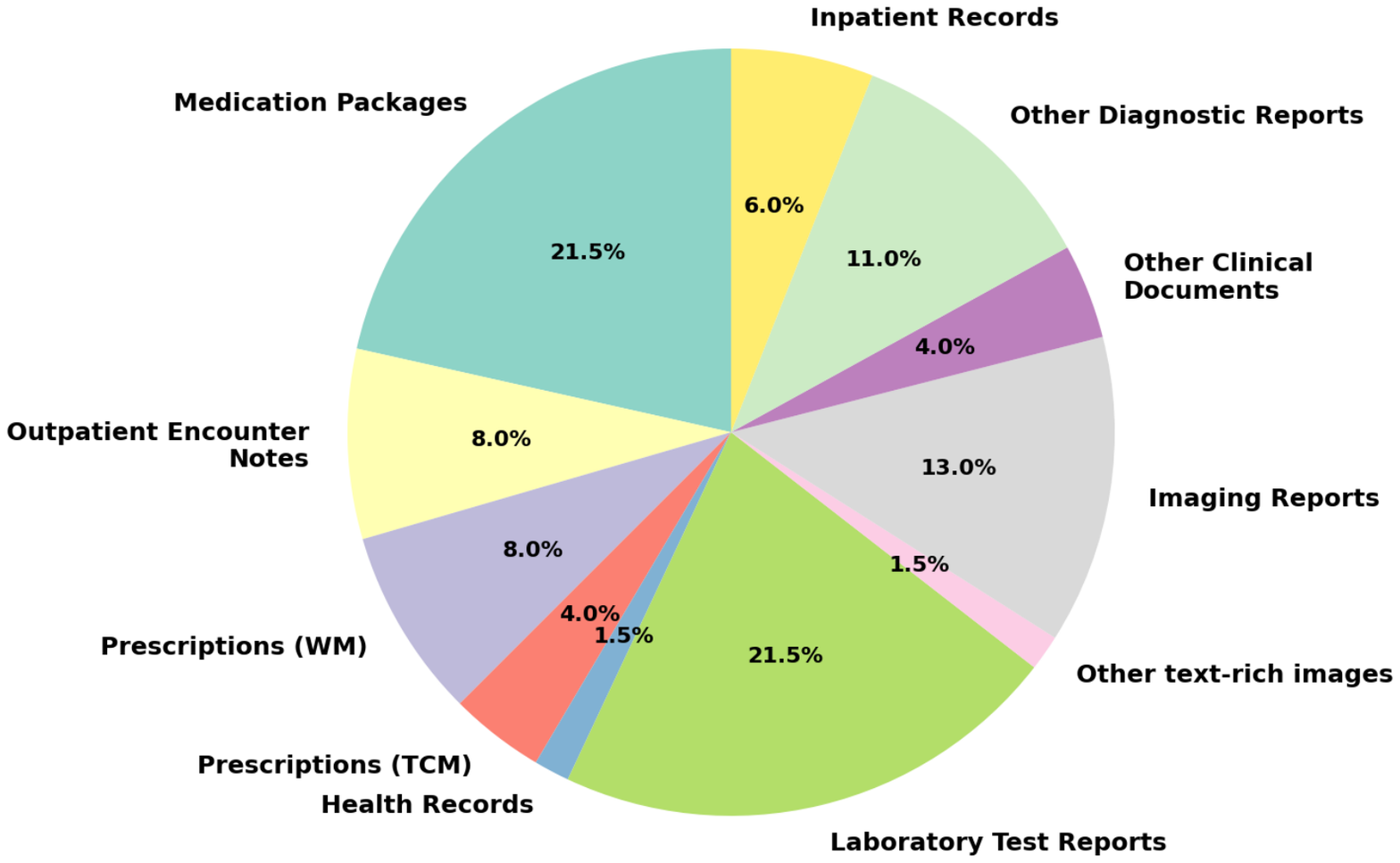}
        \caption{Distribution of MedDocBench.}
        \label{fig:meddocbench}
    \end{subfigure}
    \caption{Document Type Distribution of GMD and MedDocBench. 
    The corpus covers a broad set of document types with a long tail of less frequent categories; we downsample 
    overrepresented types to reduce bias while preserving sufficient coverage for rare forms, enabling fairer assessment of 
    cross-type robustness.}
    \label{fig:charts}
\end{figure}


We first balance the document-type distribution by downsampling overrepresented categories (Figure~\ref{fig:meddoc_dist}) 
and, as in LTR, apply rotation correction and dewarping. A VLM then generates extraction questions of two kinds—simple 
(single-fact, sampled from five question candidates) and complex (multi-fact and clinically salient). 
Each question is answered independently by two state-of-the-art VLMs, and an LLM retains only answers on which they agree.
Finally, we perform hard-example mining using the same protocol as LTR (Consensus-Before vs.\ Consensus-After).

\begin{table}[t]
\centering
\caption{Dataset Distribution of Medical Document Understanding across Document Domains, Tasks, and Consensus Buckets. 
Counts reflect unique QA instances. LTR full parsing targets standardized Markdown tables, while Simple/Complex QA probe 
single facts and abnormality/compositional reasoning. ``Hard'' and ``Easy'' correspond to Consensus-After and 
Consensus-Before, respectively. The benchmark totals 500 items across tasks.}
\label{tab:meddoc-data-dist}
\resizebox{0.98\textwidth}{!}{
\begin{tabular}{lllrr}
    \toprule
    \textbf{Domain} & \textbf{Task} & \textbf{Split} & \textbf{\# Training Samples} & \textbf{\# Benchmark Samples} \\
    \midrule
    \multirow{3}{*}{Laboratory Test Reports} 
    & Full-parse        & Hard / Easy / All & 23{,}085 \; / \; 74{,}897 \; / \; 97{,}982 & 100 \; / \; -- \; / \; 100 \\
    & Simple QA         & Hard / Easy / All & 44{,}786 \; / \; 145{,}998 \; / \; 190{,}784 & 200 \; / \; -- \; / \; 200 \\
    & Complex QA        & Hard / Easy / All & 23{,}062 \; / \; 74{,}875 \; / \; 97{,}937 & 100 \; / \; -- \; / \; 100 \\
    \midrule
    \multirow{2}{*}{General Medical Documents} 
    & Simple QA         & Hard / Easy / All & 83{,}197 \; / \; 21{,}453 \; / \; 104{,}650 & 100 \; / \; -- \; / \; 100 \\
    & Complex QA        & Hard / Easy / All & 47{,}881 \; / \; 21{,}842 \; / \; 69{,}723 & 100 \; / \; -- \; / \; 100 \\
    \midrule
    \multicolumn{3}{l}{\textbf{Total}} & \textbf{561{,}076} & \textbf{500} \\
    \bottomrule
\end{tabular}
}
\end{table}

\subsubsection{Medical Image Detection and Segmentation}

\begin{figure}[t!]
  \centering
  \includegraphics[width=0.98\textwidth]{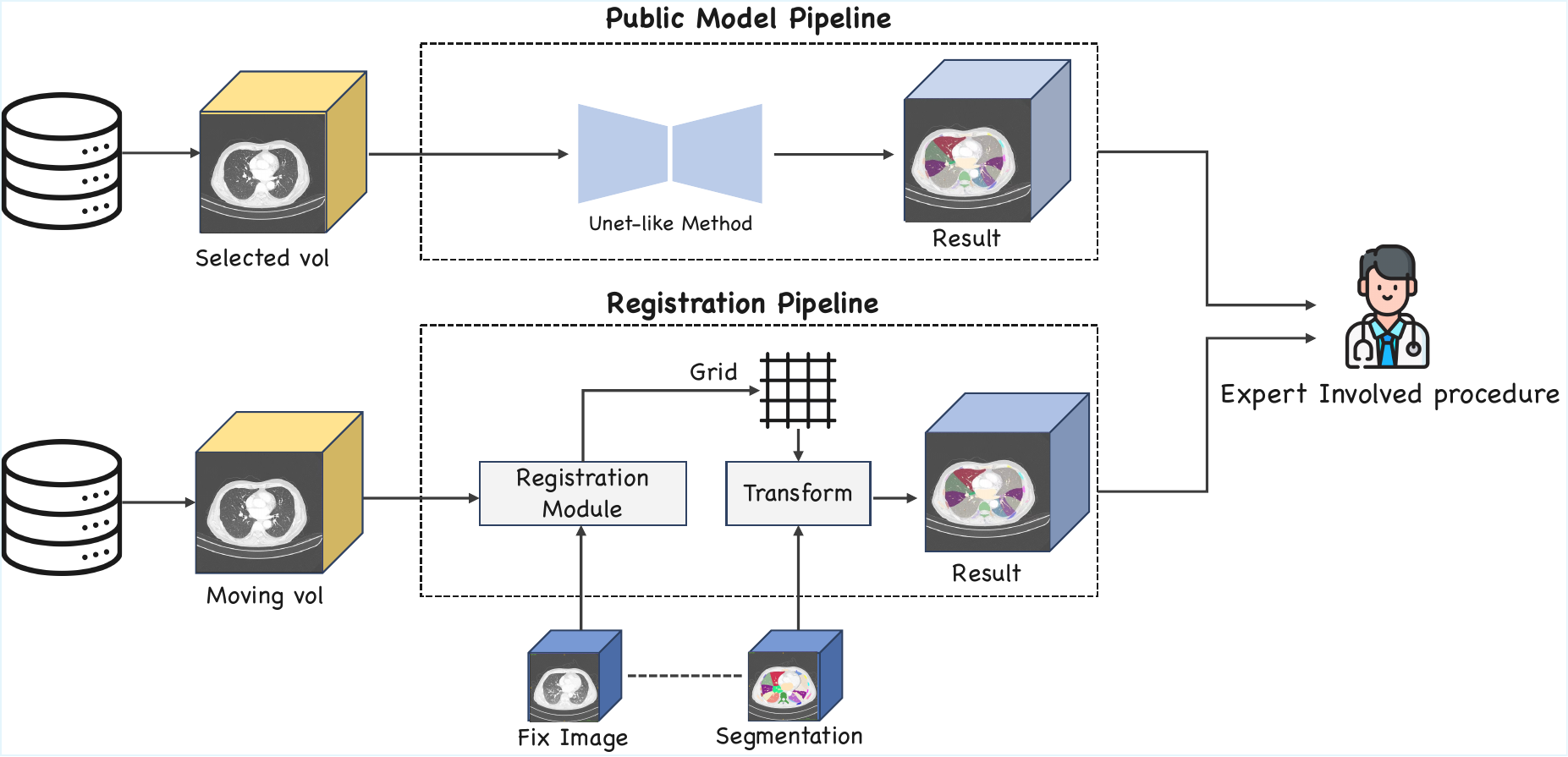}
  \caption{Overview of the Segmentation Data Generation Pipeline. Two branches produce training masks with minimal manual effort: 
  (i) public-model auto-segmentation (e.g., nnU-Net, SegFormer3D) with tiered expert review, and (ii) registration-based label transfer 
  (VoxelMorph with variational refinement) that warps template labels to target scans. Outputs receive light radiologist corrections and 
  are fed back for fine-tuning, yielding scalable, clinically reliable annotations—including rare anatomy—at a fraction of traditional 
  labeling cost.}
  \label{fig:seg_gen_pipeline}
\end{figure}
Labeling medical images at the pixel level is a labor-intensive and costly endeavor, often requiring the expertise of trained radiologists. To address this challenge, we utilizes public deep learning models and the ability of registration alogrithm. 

\textbf{Public-model pipeline:}  
Off-the-shelf nnU-Net and SegFormer3D checkpoints, pre-trained on large public corpora, generate initial segmentation masks for normal anatomy and common pathologies. A lightweight, tiered review protocol is applied: a junior annotator discards gross failures, a senior radiologist approves or swiftly corrects the remainder, and the refined masks are periodically fed back for model fine-tuning. This human-in-the-loop cycle increases reliable segmentation data by an order of magnitude within days.

\textbf{Registration pipeline:}  
We employ a VoxelMorph-based registration engine\cite{balakrishnan2019voxelmorph} augmented with classical variational refinement to estimate a dense deformation field between a fixed template and a moving CT. The resulting warp propagates the template segmentation to the moving space, yielding anatomy-agnostic, training-ready pseudo-labels for rare malformations, site-specific implants, or congenital variants in seconds Figure~\ref{fig:seg_gen_pipeline}. After propagation, a radiologist performs light post-labeling corrections, achieving clinical-grade precision while dramatically reducing manual effort.

\textbf{Impact on manual annotation:}  
By synergising public-model auto-segmentation with registration-driven label transfer, we substitute labor-intensive contouring with rapid machine drafts plus concise expert review. The former delivers high-quality masks for common structures at scale; the latter extends coverage to uncommon or scanner-specific classes without additional manual slices. Annotators thus focus on light refinement rather than delineation from scratch, slashing pure manual workload ten-fold and compressing dataset expansion from months to days.

\subsubsection{Multimodal Chain-of-Thought Reasoning}
\label{sec:datasynthesis_cot}
\begin{figure}[!t]
  \centering
  \includegraphics[width=\textwidth]{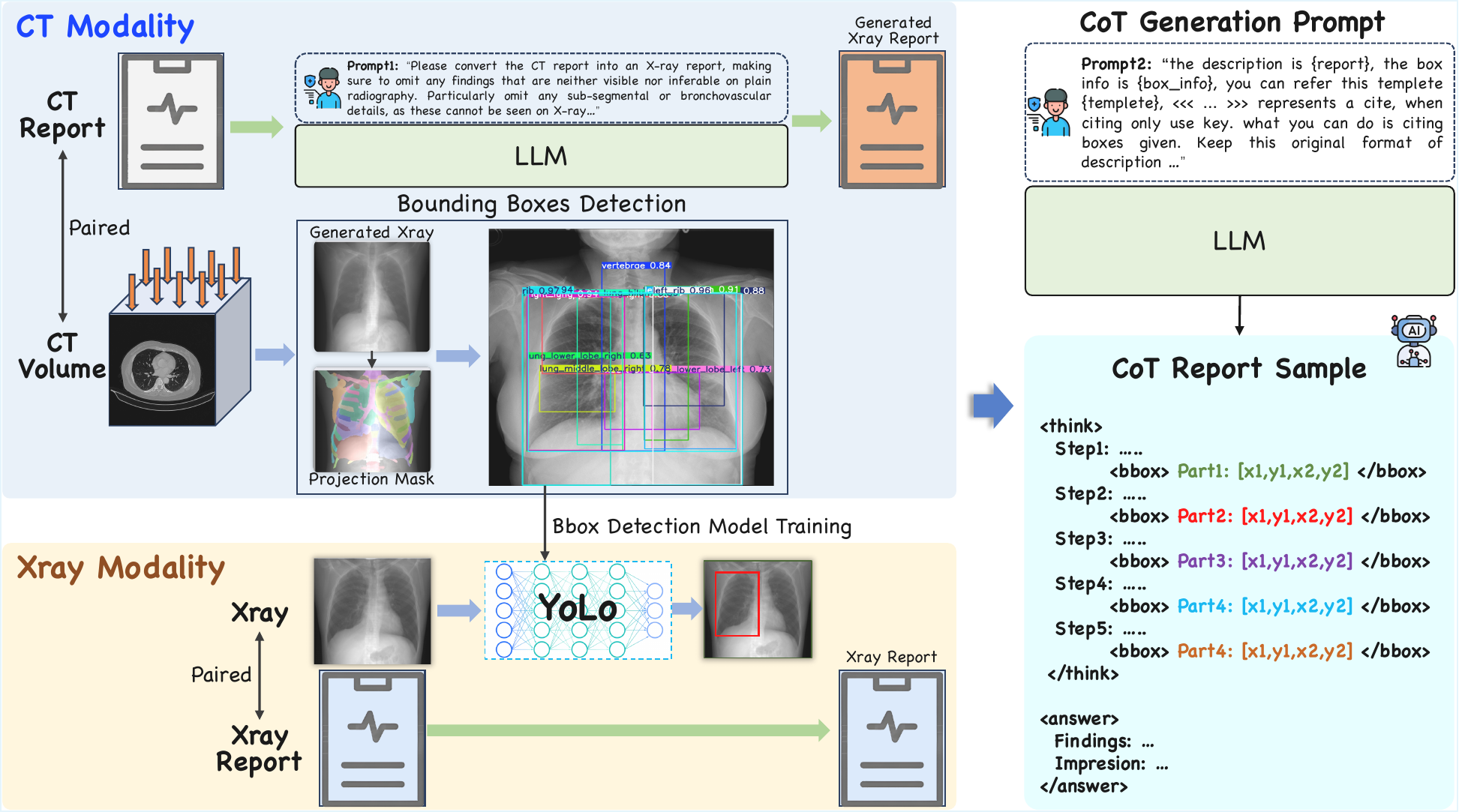}
  \caption{Overview of the CoT-generation Pipeline. CT projection is performed first to synthesize the X-ray image and its bounding boxes; these boxes are directly used to generate the CT-CoT report, while the X-ray-CoT report is produced by a detection model trained with the same projected boxes.}
  \label{fig:ct2xray}
\end{figure}

Chain-of-thought reasoning for medical images remains largely uncharted territory. Our survey shows that public datasets rarely provide step-by-step diagnostic rationales; when CoT-style captions do exist, they seldom reflect the radiologist’s true mental itinerary. Clinicians do not jump from pixels to verdict. Instead, they read the image by focusing on anatomy first: each structure is located, its morphology is measured, and its spatial relationships are evaluated. Only then is a conclusion drawn. In chest radiography, for instance, atelectasis is not declared until the “expansibility” of the two lungs is compared—lung areas, costophrenic angles, and lower border heights are appraised side by side. This deliberate, comparative scrutiny is still missing from today’s CoT annotations.

Motivated by this observation, we introduce a dataset whose diagnostic rationales are explicitly tied to a temporally ordered set of detection boxes, each corresponding to an anatomical structure examined by the radiologist. Rather than curating a fixed list of pathological regions, we adopt a comprehensive approach, modeling the full spectrum of normal, variant, and abnormal anatomical entities. A key impediment to creating such datasets is the challenge of obtaining precise, millimeter-level bounding boxes for anatomical structures in projection radiographs—a task that is both labor-intensive and prone to interobserver variability. To circumvent this, we leverage CT-derived reconstructions, where anatomical boundaries are explicitly defined, and project these labels into the 2D plane to generate accurate, reproducible bounding annotations—without additional manual effort. 

\textbf{CT Projection and Bounding-box Generation}: 
Grounded in the shared physics of CT and X-ray acquisition, we synthesise radiographs by reversible integral projection. Exploiting the Beer–Lambert law, a single 3-D CT volume is cast into line-integral projections at arbitrary angles, yielding digitally reconstructed radiographs (DRRs) that replicate the beam-attenuation, grayscale and contrast behaviour of real radiographs acquired on any X-ray equipment—without the hallucinations or domain-shift risks of AI-driven image synthesis.

Once the CT is projected, the accompanying segmentation mask is rendered at the identical geometry, producing bounding boxes that are spatially exact and free from inter-observer variability. Structures that are invisible, overlapped or occluded on the DRR remain precisely localised in the transformed mask, giving supervision that is inherently more accurate than any manually drawn annotation.

\textbf{CT Modality Report Generation}:
Before the final report is emitted, two preparatory stages are executed (Figure \ref{fig:ct2xray}). First, Prompt 1 rewrites the original CT report into an X-ray-aware narrative, stripping bronchovascular or sub-segmental details invisible on plain films while retaining clinically observable statements. Second, the draft is aligned with the projected boxes: each sentence is linked to its colour-coded bounding region (green / red / blue), grounding every finding spatially on the DRR. The LLM then produces the CoT report, sequentially citing the responsible boxes and yielding an interpretable, traceable X-ray description.

\textbf{Xray Modality Report Generation}:
A detection model is first trained on the projected DRRs and their boxes. After convergence, it is applied to real clinical X-rays to predict bounding boxes, which are fed into the identical CoT pipeline to generate the final X-ray CoT report—closing the loop from 3-D CT to 2-D radiograph reasoning.

\begin{figure}[!t]
  \centering
    \includegraphics[width=\textwidth]{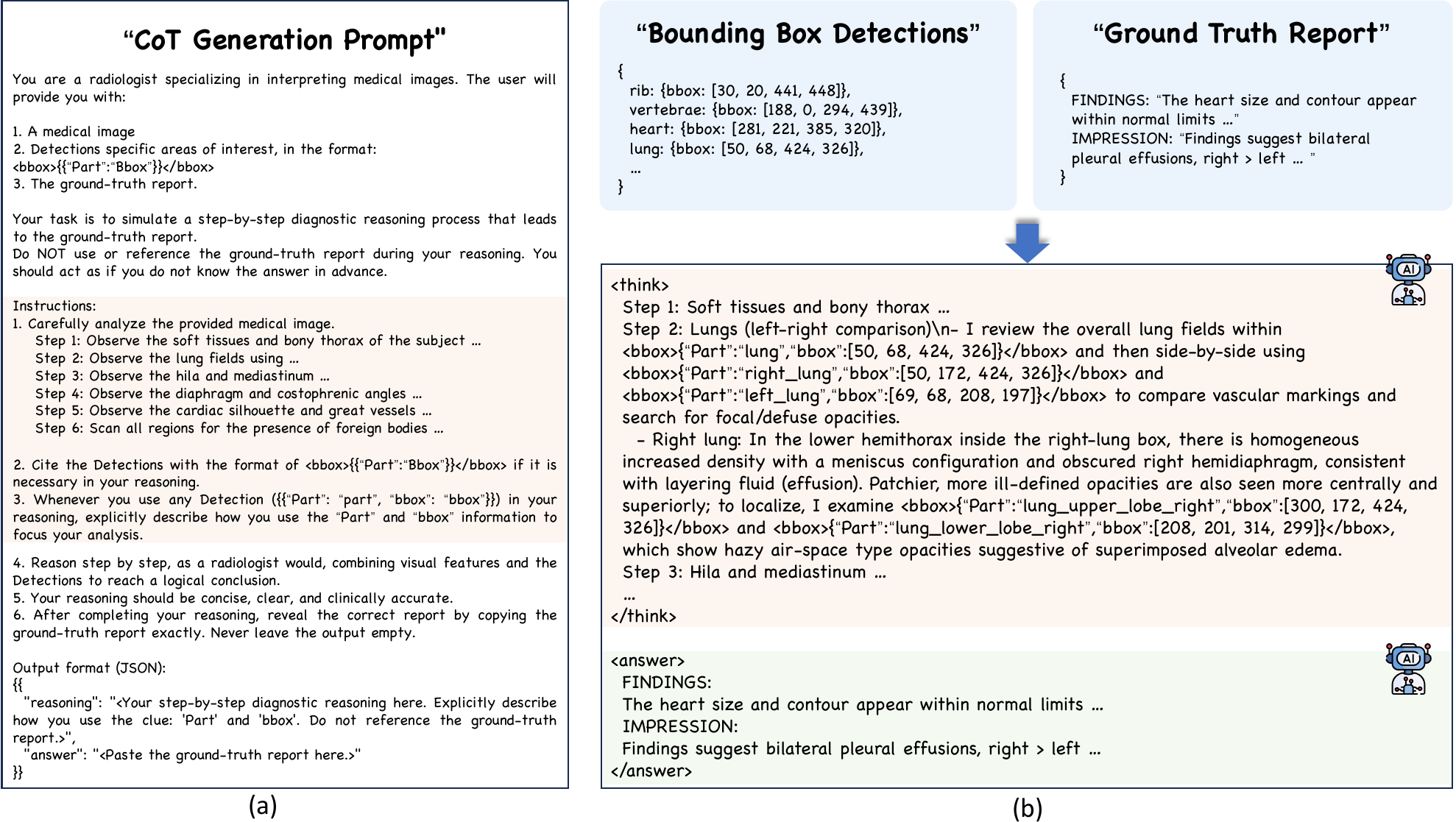}
  \caption{The Construction of Chain-of-Thought Data for Report Generation. 
  By combining the results of Bbox detection with the ground truth reports, 
  we generate CoT reasoning data using the CoT Generation Prompt.}
  \label{fig:cot_prompt}
\end{figure}

\textbf{CoT Data Generation}:
Based on the obtained detection boxes and the ground truth radiology reports, we subsequently proceed to generate Chain-of-Thought (CoT) data. GPT-5 was selected for data generation. Specifically, the prompt used for constructing CoT data is shown in the Figure~\ref{fig:cot_prompt}a. Compared to previous approaches, our method introduces two major innovations: (1) We incorporate the step of radiologists reviewing the images into the prompt (marked in red color); (2) We prompt the large model to reference the detected regions during its reasoning process. As a result, the CoT training data we construct is illustrated in the Figure~\ref{fig:cot_prompt}b. Generally, each sample consists of \texttt{<think>} and \texttt{<answer>} stages. In the \texttt{<think>} stage, the reasoning steps are organized sequentially, for example, `rib’ $\rightarrow$ `lung' $\rightarrow$ 
$\cdots$ $\rightarrow$ `heart', and at each step, the relevant detection regions \texttt{<bbox>} that facilitate the model’s reasoning are displayed.

We deliberately pursue both generation routes because these two datasets are intrinsically complementary. On the one hand, CT-derived reports provide a “God’s-eye” view: they capture details that are either invisible or easily confounded on plain radiography. For example, CT can confidently distinguish tuberculosis from bacterial pneumonia, whereas the two often appear nearly identical on X-ray. Training the model on these CT-based descriptions compels it to discover subtle X-ray patterns that correlate with otherwise occult CT findings, thereby curbing hallucination by anchoring the language to a verifiable source. Moreover, the bounding boxes derived from CT projections are essentially noise-free and offer significantly higher spatial fidelity than any box a detection model could produce on real X-rays—an additional layer of supervisory quality.

On the other hand, a curriculum composed solely of synthetic reports would never instill the terse section order, preferred negation style, or curt impression phrasing that human radiologists routinely employ. By blending in authentic X-ray reports, we preserve canonical format, domain-specific vocabulary, and clinical cadence, ensuring that the final CoT output reads like a radiologist’s note rather than a verbose CT abstract forcibly pasted onto a chest film.

\subsection{Data Statistics}

\begin{figure}[t!]
  \centering
    \includegraphics[width=0.8\textwidth]{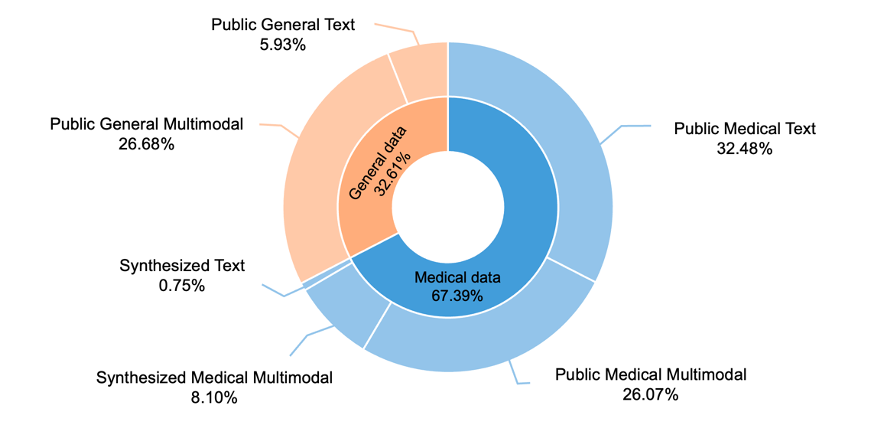}
  \caption{Type Distribution of All Training Data. The corpus combines 16.8M public and 1.6M synthesized samples produced via a standardized 
  pipeline and screened by rigorous QC (systematic deduplication and cleaning). In total it contains >12.6M images and ~3B text tokens. 
  The figure shows the proportion of each data type.}
  \label{fig:type}
\end{figure}

The training dataset comprises public data and synthesized data generated through a standardized data production pipeline. All data underwent rigorous quality control procedures, including systematic deduplication and comprehensive cleaning. Upon completion of these processes, 16.8M public data and 1.6M synthesized data are curated for model training.
Across all data, the total text token length amounts to 3B, and the total number of images exceeds 12.6M. The distribution of the training data is shown in Figure \ref{fig:type}.

\section{Model Training}

\subsection{Model Structure}

\begin{figure}[t!]
    \centering
    \includegraphics[width=0.98\linewidth]{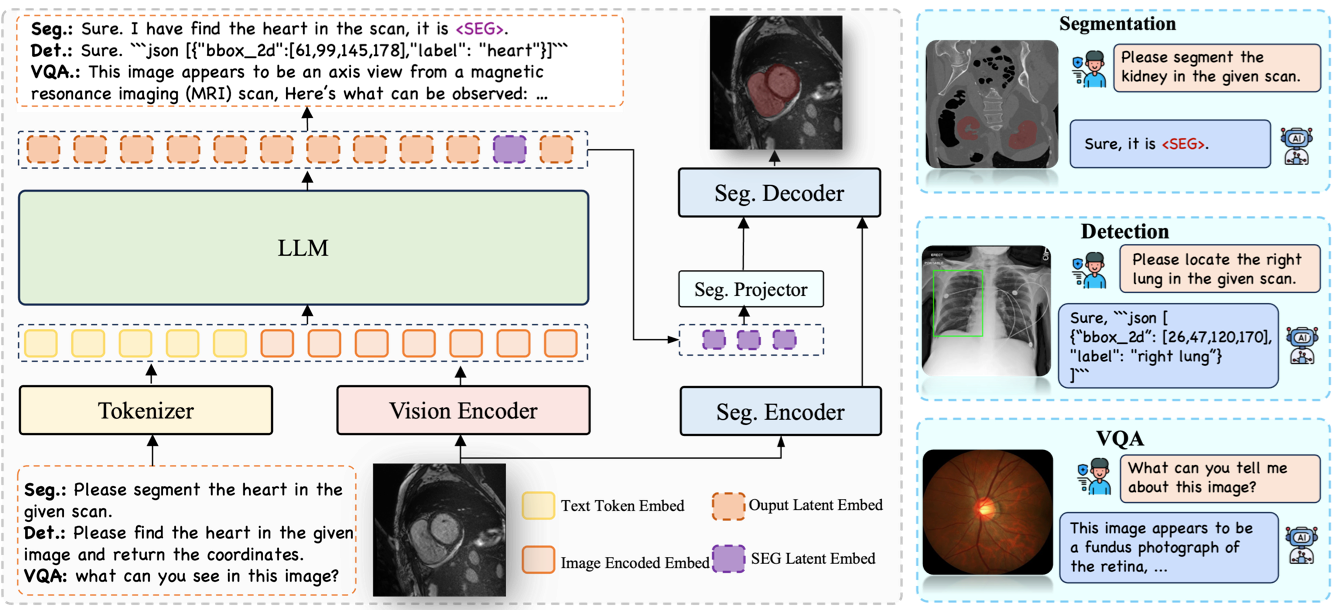}
    \caption{Model Architecture of Citrus-V. The framework consists of three components: (1) an MLLM—including the LLM, tokenizer, and a vision encoder—for high-level visual–textual 
    reasoning such as report generation, VQA, and grounding; (2) a segmentation projector that maps the "[SEG]" token produced by the MLLM into latent segmentation prompts; and (3) a 
    segmentation model that decodes the latent segmentation prompts together with semantic image features into pixel-level masks. Separate image encoders are employed to decouple 
    low-level details for segmentation from high-level semantics for other tasks, ensuring both types of tasks are optimized without semantic conflict.}
    \label{fig:citrus-V-architecture}
\end{figure}

Figure \ref{fig:citrus-V-architecture} illustrates the overall architecture of our proposed Citrus-V, which consists of three key components: a multimodal large language model (MLLM) 
for high-level visual–textual reasoning, a segmentation projector that bridges the MLLM and the segmentation model, and a segmentation model for pixel-level delineation. To seamlessly 
incorporate segmentation capabilities while preserving the reasoning power of the MLLM, we also design a unified segmentation integration strategy, which allows the model to perform 
vision-language understanding and fine-grained segmentation in a coherent, end-to-end framework.

\paragraph{MLLM for Visual-Textual Understanding} At the core of Citrus-V lies an MLLM, which serves as the reasoning backbone for tasks such as report generation, VQA, document understanding, 
and grounding. The MLLM integrates three components: (1) a vision encoder, (2) a text tokenizer, and (3) an LLM. The vision encoder transforms medical images into high-level feature tokens, 
while the tokenizer converts textual instructions into language tokens. These tokens are jointly fed into the LLM, which performs cross-modal alignment and reasoning, ultimately generating 
structured or free-text outputs depending on the task.

\paragraph{Segmentation Projector for Latent Prompt Generation} While the MLLM effectively interprets and reasons over multimodal inputs, it does not directly produce dense segmentation masks. 
To enable this capability, Citrus-V introduces a segmentation projector. Specifically, a special token, "[SEG]", is appended to the input sequence, guiding the MLLM to encode the segmentation 
intent within its hidden states. The projector then maps the hidden representation of the "[SEG]" token into a structured segmentation prompt. This prompt functions as an interface between the
MLLM and the downstream segmentation model. By isolating this projection step, Citrus-V decouples language-driven reasoning from pixel-level prediction, ensuring modularity and making the 
framework extensible to additional vision-centric tasks.

\paragraph{Segmentation Model for Pixel-Level Delineation} Similar to existing works, the segmentation model in Citrus-V is inspired by recent prompt-driven segmentation frameworks (e.g., SAM),
 but is adapted for medical imaging scenarios. It comprises three main components: a vision encoder, a prompt encoder, and a mask decoder. The vision encoder captures multi-scale image features 
 required for accurate delineation of anatomical structures. The prompt encoder processes the projected "[SEG]" prompts from the MLLM, embedding the task-specific instructions into a form compatible 
 with the mask decoder. The mask decoder combines visual features with segmentation prompts to generate pixel-level masks that delineate the target region.

Unlike conventional SAM, which primarily relies on point- or box-based visual prompts, Citrus-V leverages text-derived prompts from the MLLM. This design enables the segmentation model to be 
flexibly conditioned on natural language queries (e.g., “segment the left atrium” or “highlight tumor boundaries”), allowing for text-guided medical segmentation.

\paragraph{Dual Image Encoder Design} By coupling high-level reasoning from the MLLM with fine-grained predictions from the segmentation model, Citrus-V unifies multimodal understanding and 
pixel-level delineation within a single framework. Citrus-V employs separate image encoders: one designed for high-level semantics (supporting reasoning-oriented tasks such as VQA and grounding), 
and another specialized for low-level pixel fidelity (supporting segmentation). This separation addresses the conflicting requirements of semantic abstraction versus spatial precision, ensuring 
that both reasoning and delineation tasks are optimized without compromise.

\paragraph{Unified Segmentation Integration} We adopt Qwen2.5-VL\cite{bai2025qwen2} as the core multimodal language model, leveraging its strong visual-linguistic reasoning capability. 
For the segmentation backbone, we integrate SAM2\cite{ravi2024sam}, which provides high-quality mask generation. To seamlessly connect these components, we design a two-layer MLP segmentation 
projector, mapping MLLM features into the segmentation space.

To preserve the learned knowledge of the strong pretrained MLLM, existing frameworks adopt LoRA\cite{hu2022lora} to inject segmentation capability into the MLLM while freezing the vision 
encoder or the whole MLLM\cite{yuan2025sa2va,wu2025unibiomed}. Although LoRA is efficient, it introduces an auxiliary module outside the MLLM, thereby compromising the model’s structural 
integrity and limiting flexibility during deployment.

To address this, we design a unified segmentation integration strategy that eliminates the need for LoRA while still preserving the original reasoning ability of the MLLM. Concretely, 
in the first stage, we use large-scale text data to enable the MLLM to learn the usage of the special "[SEG]" token, which encodes segmentation intent within its hidden states. In the second stage, 
we freeze the entire MLLM and only fine-tune the segmentation projector and the entire segmentation model. This design ensures that the pretrained reasoning and instruction-following capabilities of 
the MLLM remain intact, while segmentation-specific modules are optimized for precise pixel-level delineation. Through this strategy, Citrus-V achieves a better balance between retaining multimodal 
generality and enhancing domain-specific segmentation performance, without introducing extra adapters or sacrificing model coherence.

\subsection{Training Stages}
\label{sec:TrainStage}






\begin{figure}[t!]
    \centering
    \includegraphics[width=1.0\linewidth]{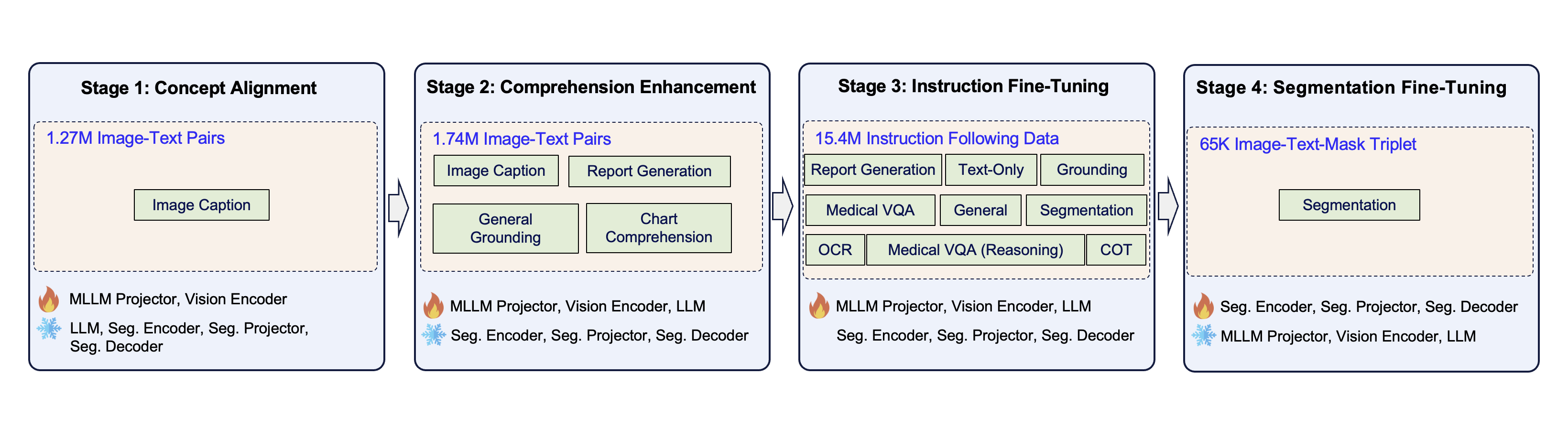}
    \caption{Four Training Stages of the Citrus-V. Consisting of concept alignment, comprehension enhancement, instruction fine-tuning, and segmentation fine-tuning }
    \label{fig:train_stages}
\end{figure}

Figure \ref{fig:train_stages} illustrates the four training stages of Citrus-V, including concept alignment, comprehension enhancement, instruction fine-tuning, and segmentation fine-tuning. The training is designed to progressively adapt the MLLM while mitigating negative transfer between multimodal reasoning and pixel-level segmentation.

\paragraph{Concept Alignment} In this stage, most model parameters are frozen, and only the MLLM projector together with the vision encoder are updated. Training primarily relies on image-caption pairs, which establish a stable mapping from visual features into the language space without disrupting the pretrained reasoning ability of the LLM. 
This step provides a lightweight but essential initialization for subsequent multimodal alignment.

\paragraph{Comprehension Enhancement} All MLLM parameters, including the projector, vision encoder, and LLM, are trainable, while the segmentation projector, encoder, and decoder remain frozen. Training incorporates a broader range of tasks, reporting detailed and structured, interpreting of medical image are preserved to establish correlations between visual features, medical concept, radiological findings and imaging diagnostic. Additionally, scientific document comprehension data such as chart, diagram, are taken considering that explication and interpretation of scientific illustrations and graphs is necessary facing with clinical documentation, laboratory reports, diagnostic image annotations and radiographic markers. 
This stage strengthens the MLLM’s multimodal comprehension capacity, while freezing the segmentation modules prevents premature interference from segmentation supervision.

\paragraph{Instruction Fine-Tuning} In the instruction fine-tuning stage, the MLLM is trained on the most diverse instruction-following data, including report generation, text-only instructions, OCR, grounding, 
medical VQA, reasoning-based chain-of-thought tasks, and segmentation instructions. It is worth noting that our empirical results 
show that directly combining segmentation loss with other tasks can damage VQA performance significantly. To this end, all MLLM parameters and segmentation modules are updated, while the gradient of segmentation modules is scaled by 0.001 with a hook function. 
In this way, for segmentation samples, the supervision is restricted to textual outputs containing the special "[SEG]" token, and applying a small mask-level loss. This design allows the MLLM to acquire the 
discourse patterns needed for segmentation queries and encode segmentation intent into the hidden state of the "[SEG]" token without losing other tasks' performance. 

\paragraph{Segmentation Fine-tuning} At this stage, all MLLM parameters are frozen, and optimization focuses exclusively on segmentation components, including the segmentation projector, the segmentation 
encoder, and the segmentation decoder. Unlike prior methods that freeze most of the segmentation backbone, the full SAM2 architecture, including the vision encoder, prompt encoder, and mask decoder, 
is fine-tuned to adapt effectively to medical imaging. With the MLLM frozen, training is computationally efficient, while full adaptation of SAM2 ensures precise and domain-specific pixel-level segmentation.

\begin{figure}[t!]
  \centering
    \includegraphics[width=\textwidth]{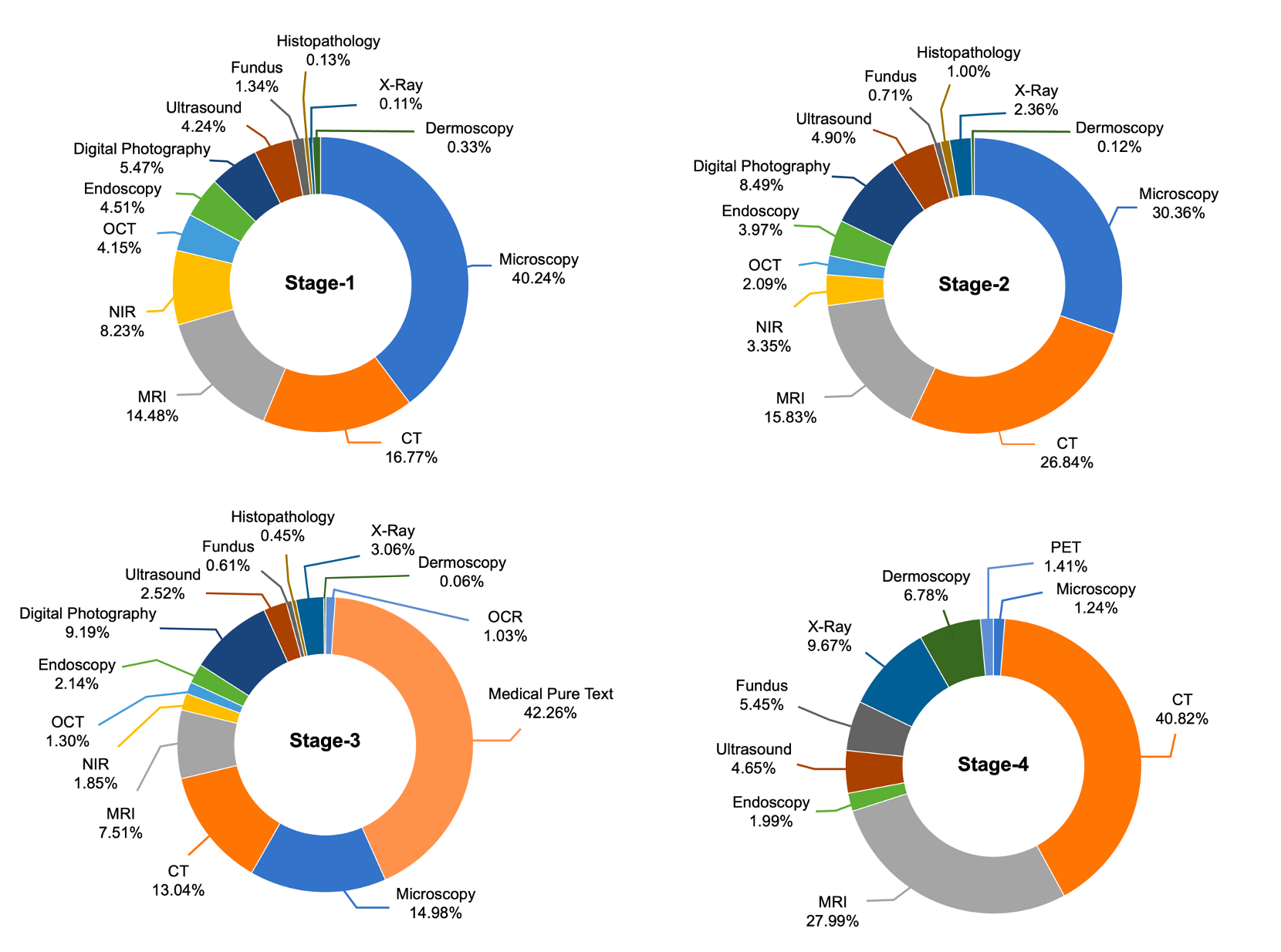}
  \caption{Modality Distribution of Medical Images in Four Training Stages. In total, there are more than 14 different imaging modalities.}
  \label{fig:modality}
\end{figure}

The training is conducted in 4 stages with various data. In Stage-1, only public medical multimodal data are used, comprising 1.27M samples. Stage-2 employs both public general and medical multimodal data, with 1.74M samples. For Stage-3, we conduct on a large-scale dataset with 15.4M samples. Stage-4 utilized 65,000 medical multimodal samples for image segmentation training. Within these data, the medical imaging data encompasses over 14 different imaging modalities.
The substantial diversity of the training data enhances the robustness and generalization capability of our model. The distributions of medical data modalities within our dataset are illustrated in Figure \ref{fig:modality}.

\subsection{Implementation Details}

\subsubsection{Training Settings}
\paragraph{Implementation Details}In the first three stages, we mainly adopt the token loss, implemented using the standard cross-entropy loss:
\begin{equation}
\mathcal{L}_{\text{token}} = - \sum_{i=1}^{N} y_i \log \hat{y}_i ,
\end{equation}
where $y_i$ denotes the ground-truth label and $\hat{y}_i$ denotes the predicted probability of the $i$-th token.

In the segmentation fine-tuning stage, in addition to the token loss, we incorporate Dice loss and binary cross-entropy (BCE) loss to enhance segmentation accuracy. The Dice loss is defined as:
\begin{equation}
\mathcal{L}_{\text{dice}} = 1 - \frac{2 \sum_{i=1}^{N} p_i g_i}{\sum_{i=1}^{N} p_i + \sum_{i=1}^{N} g_i + \epsilon},
\end{equation}
where $p_i$ and $g_i$ denote the predicted probability and ground-truth label of pixel $i$, respectively, and $\epsilon$ is a smoothing term.

The BCE loss is formulated as:
\begin{equation}
\mathcal{L}_{\text{bce}} = - \frac{1}{N} \sum_{i=1}^{N} \left[ g_i \log p_i + (1 - g_i) \log (1 - p_i) \right].
\end{equation}

The overall segmentation loss is a weighted combination of the three losses:
\begin{equation}
\mathcal{L}_{\text{seg}} = \lambda_{1} \mathcal{L}_{\text{token}} + \lambda_{2} \mathcal{L}_{\text{dice}} + \lambda_{3} \mathcal{L}_{\text{bce}},
\end{equation}
where $\lambda_{1}, \lambda_{2}, \lambda_{3}$ are balancing coefficients. In our work, the $\lambda_{1}, \lambda_{2}, \lambda_{3}$ are set 1, 2, and 1, respectively.\\

\paragraph{Training hyperparameters}
We use DeepSpeed ZeRO-2 and Accelerate to train the LLM, with AdamW as the optimizer.
The hyperparameters for this experiment were set as follows: the learning rate was initialized to 1e-5, with a separate learning rate of 2e-6 for the Vision Transformer component and 1e-5 for the aligner module. A cosine learning rate scheduler was employed. The hyperparameter settings for our model training are shown in Table \ref{tab:hyperparameters}. 
\begin{table}[t!]
    \centering
    \caption{Model Training Hyperparameters. Training uses DeepSpeed ZeRO-2 with Accelerate, AdamW, and a cosine learning-rate decay. 
    We adopt a staged schedule: stage 1 warms up the visual backbone (ViT) and the visual–text aligner while the LLM is frozen; 
    stages 2–3 jointly fine-tune the LLM with a small LR while continuing to tune ViT and the aligner; 
    stage 4 trains only the segmentation head. A dash (–) indicates the component is frozen in that stage.}
    \label{tab:hyperparameters}    

    \resizebox{0.65\textwidth}{!}{%
    \begin{tabular}{cccccc}
    \toprule
    \textbf{Train Phase} & \textbf{LLM LR} & \textbf{ViT LT} & \textbf{Aligner LR} & \textbf{Seg LR} & \textbf{Epochs} \\
    \midrule
    stage1 & - & 2e-6 & 1e-5 & - & 1\\
    stage2 & 1e-5 & 2e-6 & 1e-5 & - & 2\\
    stage3 & 1e-5 & 2e-6 & 1e-5 & - & 3\\
    stage4 & - & -  & - & 8e-5 & 10\\
    \bottomrule
    \end{tabular}
    }
    \end{table}

\subsubsection{Training Framework}

\begin{figure}[t!]
  \centering
  \includegraphics[width=1.0\textwidth]{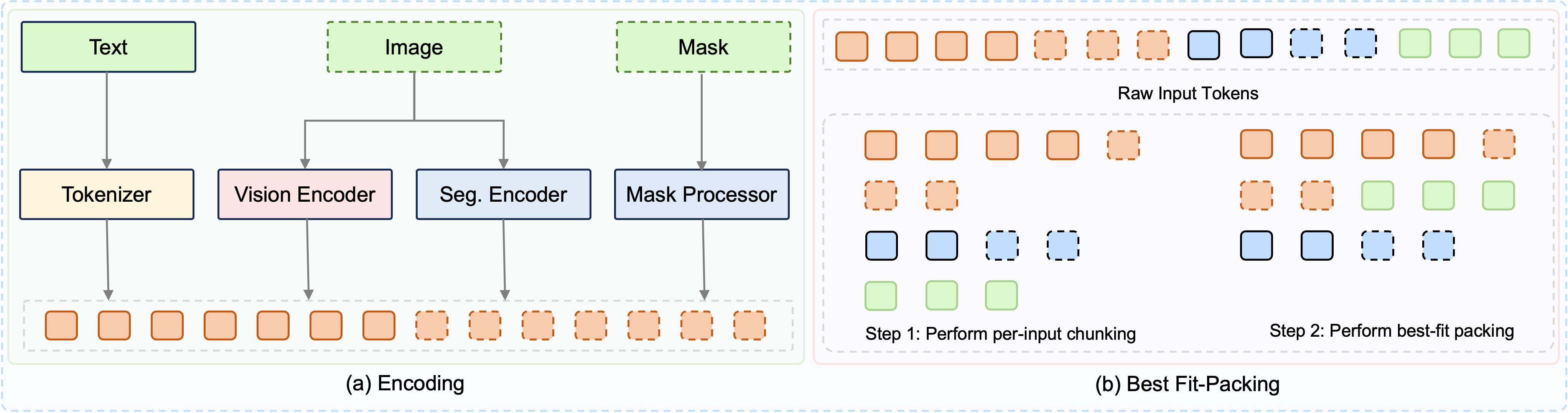}
  \caption{Best-fit Packing Strategy Utilized in Citrus-V. (a) Encoding stage. The input text and image are transformed into tokens using their respective tokenizer or encoder. 
  For segmentation data, an additional mask is processed through the mask encoder. Dashed boxes denote the encoded image or/and mask tokens. (b) Best-fit packing strategy\cite{ding2024fewer}. 
  In this illustrative example, the maximum sequence length is set to 5 tokens. Each box represents a token, with contiguous boxes of the same color indicating one encoded sample. In this example, 
  three samples of lengths 7, 4, and 3 are processed. Step 1: Longer sequences (e.g., the orange one) are split into chunks not exceeding 5 tokens. Step 2: These chunks are then arranged into 
  training sequences so as to minimize the total number of sequences, without further splitting any chunk. In the end, only a single document is truncated, which is required to conform to the 
  maximum sequence length.}
  \label{fig:training_framework_packing}
\end{figure}

We adopt the SWIFT\cite{zhao2025swift} codebase as our training framework and implement several key modifications to accommodate the requirements of our Citrus-V. Following 
existing works\cite{yuan2025sa2va, wu2025unibiomed}, we extend the input template by introducing the "g\_pixel\_values" key for the segmentation image encoder and the "masks" key 
for segmentation loss computation. This enables the framework to handle dense segmentation labels alongside standard visual and textual inputs.

In addition, we augment the original Qwen2.5-VL\cite{bai2025qwen2} textual output with an additional segmentation mask, encoded in run-length encoding (RLE)\cite{golomb1966run} to 
reduce memory consumption and facilitate efficient storage. This design allows the model to simultaneously generate textual predictions and segmentation outputs, maintaining a 
unified interface for multimodal tasks.

To enhance training efficiency and support heterogeneous data types, a best-fit packing strategy\cite{ding2024fewer} is employed in Citrus-V, as illustracted in figure \ref{fig:training_framework_packing}. 
Specifically, each input (text, image, or segmentation mask) is first tokenized or encoded into a sequence of tokens. Since different samples vary in length, longer sequences are first segmented into 
fixed-size chunks that do not exceed the predefined maximum sequence length. These chunks are then packed into training sequences using a best-fit heuristic, which minimizes the total number 
of sequences while preserving chunk integrity.

We also extend the original data-packing mechanism of SWIFT to enable mixed training of vision--language (e.g., VQA) and segmentation tasks. Within a single packed item, both segmentation and 
non-segmentation samples can co-exist. To facilitate retrieval of the corresponding segmentation mask and \texttt{g\_pixel\_values} for each ``[SEG]'' token, the input tensors are augmented 
with an additional dimension during packing. The resulting packed tensors have the shape $[B, N, C, H, W]$ where $B$ denotes the batch size, $N$ is the number of images or masks in the packed 
item, and $C, H, W$ correspond to the channel, height, and width, respectively. This design ensures that segmentation data can be efficiently accessed and processed without interfering with non-segmentation samples, 
thereby improving the overall efficiency and flexibility of mixed-data training, and leading to an approximately 8$\times$ improvement in training efficiency under comparable conditions.

\section{Experiments}

We have deconstructed the underlying capabilities of medical multimodal foundation models into five major 
categories: \textbf{medical visual question answering}, \textbf{medical textual question answering}, 
\textbf{medical document understanding}, \textbf{medical image report generation}, and \textbf{medical image detection and segmentation}. 
For each capability, we constructed dedicated evaluation datasets, which include both publicly available and 
proprietary data. Please refer Section \ref{sec:benchmarks} for detail.

In Section \ref{sec:results}, We conducted comprehensive comparative experiments on the aforementioned 
evaluation sets, pitting our model against leading commercial models and open-source medical multimodal models. 
The results demonstrate the superior performance of our proposed model across various tasks.

Given the critical influence of training data volume, training strategies, and dataset composition on model 
performance, Section \ref{sec:ablation} performed detailed ablation studies focusing on these dimensions. These 
experiments are designed to isolate and understand the individual contribution of each factor to the overall 
model capabilities, providing insights for further optimization.

\subsection{Benchmarks}
\label{sec:benchmarks}

We conducted a detailed evaluation of the model on 16 public and private evaluation datasets. These datasets are organized into 5 categories and listed in detail below. Additionally, we provide a detailed description of the construction process for two of these datasets in the following sections.

\paragraph{Medical Visual Question Answering}

primarily examines the model's ability to combine medical-domain visual data (such as CT and MRI images) 
with related textual questions to automatically generate accurate answers that conform to medical professional 
logic. The main benchmark datasets for evaluation include: VQA-RAD\cite{lau2018dataset}, MedXpertQA\cite{zuo2025medxpertqa}, 
SLAKE\cite{liu2021slake}, PATH-VQA\cite{naseem2022vision}, and PMC-VQA\cite{zhang2023pmc}.

\paragraph{Medical Textual Question Answering}

mainly evaluates the model's ability to automatically generate accurate answers that comply with medical 
professional standards for users' medical-related questions (such as disease diagnosis basis, treatment plans, 
drug side effects, and explanations of medical concepts) based on text data in the medical field (such as 
electronic medical records, medical literature, diagnosis and treatment guidelines, and drug instructions). The 
datasets mainly used for evaluation include: PubMedQA\cite{jin2019pubmedqa}, MedMCQA\cite{pal2022medmcqa}, 
MedQA\cite{jin2020diseasedoespatienthave}, MedXpertQA\cite{zuo2025medxpertqa}, CMMLU\cite{li2023cmmlu}, 
Medbullets\cite{chen2025benchmarking}, and SuperGPQA\cite{du2025supergpqa}.

\paragraph{Medical Document Understanding}

examines the model's capability to perform in-depth parsing of information such as text, tables, 
and charts within real-world medical documents (e.g., lab reports, prescription slips) and accurately extract key medical 
knowledge.

\paragraph{Medical Image Report Generation}

mainly evaluates the model's ability to automatically generate structured, clinical-grade radiological 
diagnosis reports by integrating medical imaging data. The main dataset used for evaluation includes: CheXpert Plus\cite{chambon2024chexpert}.

\paragraph{Medical Image Detection and Segmentation}

primarily examines the use of large language models to achieve the detection and segmentation of anatomical 
structures in medical imaging scans. The main benchmark datasets for evaluation include: \verb+MeCOVQA-G++, 
MedSegBench\cite{kucs2024medsegbench} for segmentation task, and MedSAM2\cite{ma2025medsam2} for detection task, 
in which \verb+MeCOVQA-G++ is cleaned from MeCOVQA-G\cite{huang2025towards}, MedSegBench covers a wide range of modalities.

\subsubsection{MedDocBench}
\label{sec:MedDocBench}

We construct \textsc{MedDocBench}, a publicly available benchmark for medical document understanding covering routine, 
patient-uploaded artifacts from online consultations.

\paragraph{Benchmark Statistics.} 
\textsc{MedDocBench} comprises two tracks: Laboratory Test Reports (LTR) and General Medical Documents (GMD). 
LTR includes full parsing, simple QA, and complex QA; GMD includes simple QA and complex QA. 
We release the Hard test sets, which yield a more discriminative and realistic evaluation and better separate 
top-performing VLMs. See Section~\ref{sec:MedDocTrain} for details on the tracks, splits, and the ground-truth labeling 
pipeline, and Table~\ref{tab:meddoc-data-dist} and Figure~\ref{fig:meddocbench} for data distribution.

\paragraph{Evaluation Protocol.}
LTR: All predictions are canonicalized as a preprocessing step. For full parsing, models output a Markdown table; we perform 
bipartite matching between predicted and gold \texttt{entry\_name}s, then assess field-level correctness for \texttt{entry\_name}, 
\texttt{result}, \texttt{reference}, and \texttt{unit}. We report micro Precision/Recall/F1 at the field level and macro-averaged 
scores at the document (image) level. For complex QA, models return JSON containing the \texttt{result}, \texttt{reference}, 
and an abnormality label; evaluation follows the same matching procedure with P/R/F1 reporting. For simple QA, we compute 
exact-match accuracy after canonicalization. In all LTR subtasks, an LLM judge is used as a fallback when rule-based matching failed.
GMD: Free-form answers are scored by an LLM judge given the question, the predicted answer, and the gold answer. 
The judge produces a continuous score \(s \in [0,1]\); we report the mean score. We release the evaluation scripts to support 
reproducibility.

\subsubsection{MeCoVQA-G+}
To the best of our knowledge, datasets for training and evaluating text–segmentation alignment in the medical domain are extremely scarce. One of the few publicly available resources is MeCoVQA-G, which was recently introduced alongside the MedPlib paper\cite{chambon2024chexpert}. MeCoVQA-G is a large-scale, pixel-level VQA subset of the MeCoVQA family, constructed by pairing biomedical images with natural-language questions that explicitly ask the model to segment a given anatomical structure or lesion. Each sample contains a 2-D image slice, a templated question targeting a specific anatomical class, and the corresponding binary segmentation mask as the ground-truth answer.
The released split is 100K training pairs and 2,344 test pairs.  

However, the formidable challenge of verifying and annotating medical imagery has left the original train/test split riddled with obvious label errors, undermining the scientific rigor of both training and evaluation. To address this, we present MeCOVQA-G+, a thoroughly re-annotated and expanded edition of the corpus. Beyond simply correcting the errors, MeCOVQA-G+ increases both the scale and modality diversity of its predecessor, delivering a more reliable and comprehensive benchmark for medical text-to-segmentation tasks.

We will release the MeCOVQA-G+ test set to establish a new benchmark that specifically addresses the above issues. The set comprises 3,157 carefully curated text–segmentation pairs, in which we have corrected all evident mis-alignments and duplicates. The samples span a wide range of modalities, including X-ray, CT, MRI, ultrasound, and endoscopy. Each image has been meticulously reviewed by a team of medical experts to ensure the accuracy of the segmentation masks.

\subsection{Main Results}
\label{sec:results}

To conduct a more comprehensive evaluation of model performance, we carried out detailed testing on both commercial models and open-source models using medical datasets. For this purpose, we categorized the models under evaluation into three groups: 

\paragraph{Proprietary Models}
GPT 4.1\cite{OpenAI_GPT4.1_2024}, GPT 5\cite{OpenAI_GPT5_2025}, Doubao Seed 1.6\cite{Doubao_DoubaoSeed1.6_2024}

\paragraph{Open-Source Models (<10B)}
MedGemma 4B\cite{sellergren2025medgemma}, Qwen2.5-VL 7B\cite{bai2025qwen2}, 
HuatuoGPT-V 7B\cite{chen2024huatuogpt}, Lingshu 7B\cite{xu2025lingshu}

\paragraph{Open-Source Models (>10B)}
MedPlib 14B\cite{huang2025towards}, MedGemma 27B\cite{sellergren2025medgemma}, 
Qwen2.5-VL 32B\cite{bai2025qwen2}, Lingshu 32B\cite{xu2025lingshu}, 
HealthGPT 14B\cite{lin2025healthgpt}, HealthGPT 32B\cite{lin2025healthgpt}, 
HuatuoGPT-V 34B\cite{chen2024huatuogpt}

\subsubsection{Medical Textual Question Answering}

We conducted a comprehensive evaluation of both commercial and open-source models on open-source medical text question-answering benchmarks to assess their performance in medical text reasoning shown in Table \ref{tab:results:mtqab}. Our findings are as follows. Firstly, despite being trained on a large amount of multimodal data, our Citrus-V 33B model achieved the highest overall average performance, outperforming the second-best model, Lingshu 32B, by 1.44\%. Secondly, although MedPlib 14B possesses both image detection/segmentation and medical question-answering capabilities similar to ours, its performance on pure text-based question-answering benchmarks was relatively poor. More specifically, our model achieved the best results on 4 out of 7 test sets, including MedMCQA, CMMLU, Medbullets, and SuperGPQA, and ranked second on the remaining 3 test sets, namely PubMedQA, MedQA, and MedXpertQA.

Among models with fewer than 10 billion parameters, our Citrus-V 8B model still achieves the highest average performance across all evaluation sets, ranking first on 5 out of 7 datasets and outperforming the second-best model, Lingshu 7B, by 0.82\%.

\begin{table}[t!]
  \centering
  \caption{Performance of Models on Medical Textual Question Answering Benchmarks.
  The best results on each benchmark and average accuracy are highlighted in bold, and the scores with underline indicate the second best.}
  \label{tab:results:mtqab}
  \resizebox{\textwidth}{!}{%
  \begin{tabular}{lcccccccc}
    \toprule
    \textbf{Model} & \textbf{PubMedQA} & \textbf{MedMCQA} & \textbf{MedQA} & \textbf{MedXpertQA} & \textbf{CMMLU} & \textbf{Medbullets} & \textbf{SuperGPQA} & \textbf{Avg.} \\
    \midrule
    \multicolumn{9}{c}{Proprietary Models} \\
    \midrule
    GPT 4.1 & 76.00 & 77.07 & 87.98 & 30.82 & 81.02 & 73.38 & 50.6 & 68.12 \\
    GPT 5 & 78.00 & 62.99 & 76.96 & 40.75 & 82.93 & 87.30 & 49.54 & 68.35 \\
    Doubao Seed 1.6 & 76.00 & 75.06 & 93.48 & 30.67 & 91.67 & 76.62 & 55.19 & 71.24 \\
    \midrule
    \multicolumn{9}{c}{Open-Source Models (<10B)} \\
    \midrule
    MedGemma 4B & 73.00 & 52.26 & 55.54 & 13.1 & 43.96 & 42.53 & 21.52 & 43.13 \\
    Qwen2.5-VL 7B & \textbf{75.80} & 53.40 & 57.50 & 12.40 & 68.80 & 36.69 & 26.39 & 47.28 \\
    HuatuoGPT-V 7B & 73.60 & 51.95 & 52.95 & 10.33 & \underline{71.12} & 37.66 & 22.11 & 45.67 \\
    Lingshu 7B & \underline{75.40} & \textbf{56.13} & \underline{63.39} & \underline{16.45} & 69.02 & \underline{52.92} & \underline{27.51} & \underline{51.55} \\
    \rowcolor{green!12} Citrus-V 8B & 74.80 & \underline{55.10} & \textbf{64.89} & \textbf{16.90} & \textbf{71.19} & \textbf{54.22} & \textbf{29.47} & \textbf{52.37} \\
    \midrule
    \multicolumn{9}{c}{Open-Source Models (>10B)} \\
    \midrule
    MedPlib 14B & 49.40 & 1.63 & 7.38 & 0.45 & 15.53 & 1.30 & 0.22 & 10.84 \\
    MedGemma 27B & \textbf{79.00} & 63.23 & \textbf{81.15} & 22.01 & 60.24 & \underline{65.58} & 33.18 & 57.77 \\
    Qwen2.5-VL 32B & 68.60 & 62.71 & 71.33 & 15.88 & \underline{82.60} & 50.65 & 38.26 & 55.72 \\
    Lingshu 32B & 78.20 & \underline{65.05} & 74.94 & \textbf{22.86} & 82.37 & 63.31 & \underline{40.80} & \underline{61.08} \\
    HealthGPT 14B & 69.40 & 63.33 & 66.93 & 12.45 & 55.36 & 50.00 & 25.59 & 49.01 \\
    HealthGPT 32B & 74.20 & 64.04 & 68.89 & 13.84 & 69.47 & 46.43 & 35.43 & 53.19 \\
    HuatuoGPT-V 34B & 71.00 & 55.08 & 58.52 & 12.2 & 77.64 & 39.29 & 28.06 & 48.83 \\
    \rowcolor{blue!12} Citrus-V 33B & \underline{78.40} & \textbf{65.62} & \underline{80.28} & \underline{22.20} & \textbf{83.27} & \textbf{66.23} & \textbf{41.63} & \textbf{62.52} \\
    \bottomrule
  \end{tabular}%
  }
\end{table}

\subsubsection{Medical Visual Question Answering}

Table \ref{tab:results:mvqab} presents a detailed comparison between Lingshu and a diverse set of both 
proprietary and open-source MLLMs across seven medical multimodal benchmarks. 

Among models with larger than 10B parameters, Citrus-V 33B achieves the highest average score across the evaluation sets, outperforming Lingshu32B (the second-ranked model) by 0.59\%. It also establishes state-of-the-art (SOTA) performance on three evaluation sets, \textit{i.e.}, VQA-RAD, SLAKE, and PMC-VQA, while ranking second on the PATH-VQA test set.

Among models with fewer than 10 billion parameters, Citrus-V 8B ranks second in terms of average evaluation metrics, trailing the top model, Lingshu 7B, by 0.52\%. Additionally, it achieves state-of-the-art (SOTA) performance on two evaluation sets, namely SLAKE and PATH-VQA.

Although commercial models maintain a clear advantage on medical text QA benchmarks, in medical visual question answering (VQA) scenarios, open-source models fine-tuned on medical multimodal data surpass commercial models in terms of average performance metrics. This is primarily because commercial models excel in text reasoning capabilities. This distinction is evident in that, while commercial models underperform in certain perceptual tasks, such as VQA-RAD, SLAKE, and PATH-VQA, they demonstrate superior results in reasoning-intensive tasks like MedXpertQA.

\begin{table}[t!]
  \centering
  \caption{Performance of Models on Medical Visual Question Answering Benchmarks.
  The best results on each benchmark and average accuracy are highlighted in bold, and the scores with underline indicate the second best.}
  \label{tab:results:mvqab}
  \resizebox{0.9\textwidth}{!}{%
  \begin{tabular}{lcccccc}
    \toprule
    \textbf{Model} & \textbf{VQA-RAD} & \textbf{MedXpertQA} & \textbf{SLAKE} & \textbf{PATH-VQA} & \textbf{PMC-VQA} & \textbf{Avg.} \\
    \midrule
    \multicolumn{7}{c}{Proprietary Models} \\
    \midrule
    GPT 4.1 & 62.53 & 43.35 & 72.54 & 54.97 & 38.76 & 54.43 \\
    GPT 5 & 68.37 & 51.48 & 65.82 & 31.74 & 36.10 & 50.70 \\
    Doubao Seed 1.6 & 33.49 & 45.75 & 67.28 & 47.58 & 49.94 & 48.81 \\
    \midrule
    \multicolumn{7}{c}{Open-Source Models (<10B)} \\
    \midrule
    MedGemma 4B & \textbf{72.06} & 22.05 & 78.32 & 48.64 & 48.02 & 53.82 \\
    Qwen2.5-VL 7B & 66.30 & 20.75 & 67.86 & 42.30 & 50.86 & 49.61 \\
    HuatuoGPT-V 7B & 67.85 & 22.30 & 69.39 & 44.29 & 53.84 & 51.53 \\
    Lingshu 7B & \underline{68.74} & \textbf{26.90} & \underline{82.90} & \underline{60.23} & \textbf{55.77} & \textbf{58.91} \\
    \rowcolor{green!12} Citrus-V 8B & 64.30 & \underline{25.10} & \textbf{84.91} & \textbf{62.00} & \underline{55.64} & \underline{58.39} \\
    \midrule
    \multicolumn{7}{c}{Open-Source Models (>10B)} \\
    \midrule
    MedPlib 14B & 45.45 & - & 38.54 & 40.02 & 44.40 & - \\
    MedGemma 27B & 63.86 & \textbf{33.10} & 76.17 & 47.6 & 45.35 & 53.22 \\
    Qwen2.5-VL 32B & 72.28 & 25.30 & 76.36 & 41.58 & 53.58 & 53.82 \\
    Lingshu 32B & \underline{75.39} & \underline{31.00} & \underline{87.68} & \textbf{64.76} & \underline{57.23} & \underline{63.21} \\
    HealthGPT 14B & 64.08 & 24.55 & 67.43 & 58.67 & 56.90 & 54.33 \\
    HealthGPT 32B & 64.75 & 26.40 & 70.58 & 62.93 & 54.93 & 55.92 \\
    HuatuoGPT-V 34B & 63.64 & 22.65 & 73.02 & 44.92 & 56.79 & 52.20 \\
    \rowcolor{blue!12} Citrus-V 33B & \textbf{77.83} & 29.15 & \textbf{88.40} & \underline{63.89} & \textbf{59.74} & \textbf{63.80} \\
    \bottomrule
  \end{tabular}%
  }
\end{table}

\subsubsection{Medical Document Understanding}

To evaluate the performance of multimodal language models in medical report interpretation, we conducted assessments in two primary scenarios: laboratory test report question answering (QA) and comprehensive medical document QA. As the entire evaluation dataset is based in Chinese, the commercial model Doubao Seed 1.6 significantly outperforms the GPT series, with this advantage becoming even more pronounced in complex QA tasks.

As shown in Table \ref{tab:results:mdcb}, among all open-source models with more than 10 billion parameters, Citrus-V 33B demonstrates comprehensive leadership across all evaluation sets, outperforming Qwen2.5-VL 32B (the second-ranked model) by approximately 21.58\%. The experimental results also indicate that, although Lingshu 32B performs strongly in medical question-answering scenarios, its effectiveness in medical document understanding drops significantly compared to its base model, with a decrease of about 7.74\%. Notably, our model also surpasses the current state-of-the-art commercial models, such as Doubao Seed 1.6, in terms of average metrics.

Among models with fewer than 10 billion parameters, the Citrus-V 8B model achieves an average evaluation score of 86.2\%, outperforming the second-best model, Qwen2.5-VL 7B, by approximately 20.02\%, and still maintaining a significant lead over commercial models.

While current medical multimodal language models have achieved promising results in tasks such as medical visual QA, the evaluation shows that all models exhibit varying degrees of performance decline in rich-text image understanding tasks compared to the open-source model Qwen2.5-VL. This suggests that integrating medical image analysis and rich-text image understanding into a unified foundation model remains a significant challenge.

\begin{table}[t!]
  \centering
  \caption{Performance of Models on Medical Document Understanding Benchmarks.
  The best results on each benchmark and average accuracy are highlighted in bold, and the scores with underline indicate the second best.}
  \label{tab:results:mdcb}
  \resizebox{0.9\textwidth}{!}{%
  \begin{tabular}{lcccccc}
    \toprule
    \multirow{2}{*}{\textbf{Model}} &
    \multicolumn{3}{c}{\textbf{Laboratory Test Report}} & \multicolumn{2}{c}{\textbf{Medical Doc Report}} &
    \multirow{2}{*}{\textbf{Avg.}} \\
    \cmidrule(lr){2-6}
    & \textbf{Extract} & \textbf{Simple QA} & \textbf{Complex QA} & \textbf{Simple QA} & \textbf{Complex QA} & \\
    \midrule
    \multicolumn{7}{c}{Proprietary Models} \\
    \midrule
    GPT 4.1 & 66.86 & 71.4 & 36.66 & 53.17 & 55.04 & 56.63 \\
    GPT 5 & 69.05 & 78.60 & 56.78 & 82.92 & 64.50 & 70.37 \\
    Doubao Seed 1.6 & 80.00 & 84.72 & 70.90 & 81.00 & 70.92 & 77.51 \\
    \midrule
    \multicolumn{7}{c}{Open-Source Models (<10B)} \\
    \midrule
    MedGemma 4B & 27.61 & 19.87 & 9.56 & 29.75 & 25.92 & 22.54 \\
    Qwen2.5-VL 7B & \underline{73.23} & \underline{83.62} & \underline{31.22} & \underline{71.92} & \underline{70.92} & \underline{66.18} \\
    HuatuoGPT-V 7B & 23.10 & 10.26 & 4.56 & 25.17 & 14.21 & 15.46 \\
    Lingshu 7B & 60.06 & 60.48 & 23.45 & 53.92 & 52.28 & 50.05 \\
    \rowcolor{green!12} Citrus-V 8B & \textbf{91.21} & \textbf{97.38} & \textbf{84.28} & \textbf{79.75} & \textbf{78.38} & \textbf{86.20} \\
    \midrule
    \multicolumn{7}{c}{Open-Source Models (>10B)} \\
    \midrule
    MedPlib 14B & - & - & - & - & - & - \\
    MedGemma 27B & 26.15 & 32.97 & 5.50 & 29.50 & 22.21 & 23.27 \\
    Qwen2.5-VL 32B & \underline{71.40} & \underline{83.41} & 30.54 & \underline{73.29} & \underline{73.38} & \underline{66.40} \\
    Lingshu 32B & 63.46 & 73.36 & \underline{32.38} & 61.25 & 62.83 & 58.66 \\
    HealthGPT 14B & 22.22 & 17.90 & 3.17 & 24.67 & 16.00 & 16.79 \\
    HealthGPT 32B & 20.95 & 23.80 & 2.29 & 25.08 & 20.46 & 18.52 \\
    HuatuoGPT-V 34B & 22.78 & 8.52 & 10.88 & 22.50 & 16.12 & 16.16 \\
    \rowcolor{blue!12} Citrus-V 33B & \textbf{90.01} & \textbf{96.29} & \textbf{87.99} & \textbf{83.08} & \textbf{82.54} & \textbf{87.98} \\
    \bottomrule
  \end{tabular}%
  }
\end{table}

\subsubsection{Medical Image Report Generation}

Beyond conventional QA tasks, we further evaluate our model on the task of generating radiology reports. CheXpert Plus serves as the benchmark for this part of the experiments. For a fair comparison, we benchmark against a broad set of competitors, categorized into two groups: Proprietary Models and Open-Source Models. 

In the Proprietary Models category, well-known systems such as GPT 4.1, GPT 5, and Doubao Seed1.6 are used as baselines. For open-source models, we differentiate by parameter scale at 10B, yielding two subgroups: Open Source Models (<10B) and Open Source Models (>10B). Specifically, in the <10B group, we include MedGemma 4B, Qwen2.5-VL 7B, HuatuoGPT-V 7B, and Lingshu 7B. In the >10B group, our candidates comprise MedPlib 14B, MedGemma 27B, Qwen2.5-VL 32B, Lingshu 32B, HealthGPT 14B, HealthGPT 32B, and HuatuoGPT-V 34B.

The experimental results are shown in Table \ref{tab:results:mirgb}. We adopt three metrics, ROUGE-L, CIDEr, and RaTE, as our evaluation criteria (ROUGE-L\cite{lin2004rouge}, CIDEr\cite{vedantam2015cider}, and RaTE score(RaTE)\cite{zhao2024ratescore}). ROUGE-L and CIDEr are used to measure the quality of model-generated reports with respect to reference answers. Following ReXrank and Lingshu, we utilize RaTEScore to measure candidate reports.

Within the Open‑Source Models (>10B) group, our Citrus-V 33B achieves the best performance. In CheXpert Plus benchmark, our model achieves a large improvement over the second-place Lingshu 32B. Specifically, ROUGE‑L increases by $+4.29\%$ (from 25.29 to 29.58), CIDEr by $+31.24\%$ (from 77.42 to 108.66), and RaTE by $+6.27\%$ (from 46.18 to 52.45). 

In the Open‑Source Models (<10B) group, our Citrus-V 8B outperforms almost all competitors, achieving an average improvement of +3.82\% compared to the second-place MedGemma 4B. Specifically, ROUGE-L increases by +2.93\% (from 26.01 to 28.94), and CIDEr increases by +8.71\% (from 85.86 to 94.57). 

\begin{figure}[!t]
  \centering
    \includegraphics[width=\textwidth]{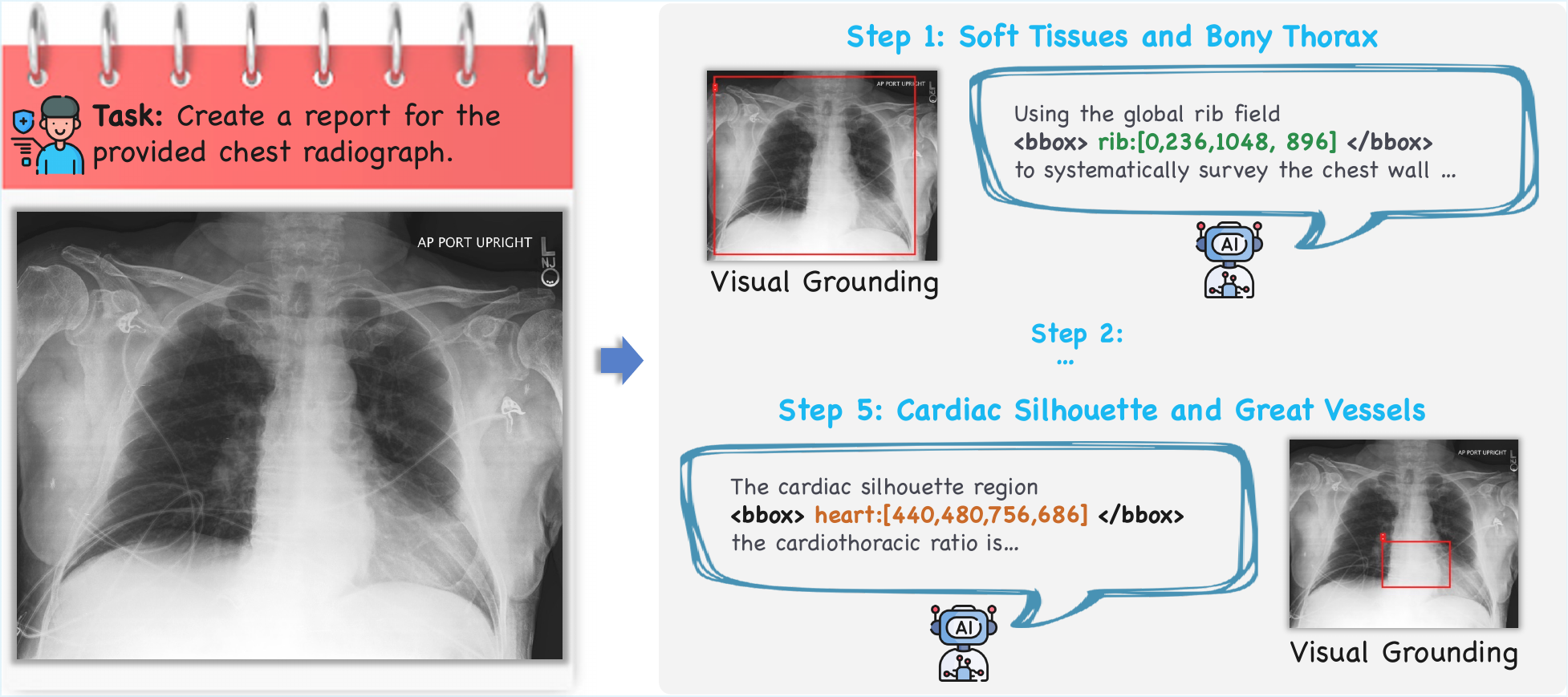}
  \caption{CoT Visualization of Citrus-V on Task of Medical Report Generation. During the reasoning process, Citrus-V leverages visual grounding to localize the corresponding anatomical structures or lesion areas within the medical images, thereby completing a rigorous reasoning process.}
  \label{fig:cot:visualization}
\end{figure}
In addition, we present visualization results in Figure~\ref{fig:cot:visualization}. As shown by the results, Citrus-V follows the workflow of radiologists when generating reports, exhibiting a rigorous reasoning process. To mitigate the occurrence of hallucinations during reasoning and to further enhance the transparency and interpretability of the model’s thought process, we trained our model to possess grounding capabilities. This enables the model to localize lesions during the reasoning process.

\begin{table}[t!]
  \centering
  \caption{Performance of Models on Medical Image Report Generation Benchmarks.
  The best results on each benchmark are highlighted in bold, and the scores with underline indicate the second best.}
  \label{tab:results:mirgb}
  \resizebox{0.5\textwidth}{!}{%
  \begin{tabular}{lccc}
    \toprule
    \multirow{2}{*}{\textbf{Model}} &
    \multicolumn{3}{c}{\textbf{CheXpert Plus}}  \\
    \cmidrule(lr){2-4}
    & \textbf{ROUGE-L} & \textbf{CIDEr} & \textbf{RaTE}  \\
    \midrule
    \multicolumn{4}{c}{Proprietary Models} \\
    \midrule
    GPT 4.1 & 24.50 & 78.80 & 45.50 \\
    GPT 5 & 24.48 & 86.46 & 51.26  \\
    Doubao Seed 1.6 & 19.27 & 61.92 & 45.49 \\
    \midrule
    \multicolumn{4}{c}{Open-Source Models (<10B)} \\
    \midrule
    MedGemma 4B & 26.01 & \underline{85.86} & \textbf{51.23} \\
    Qwen2.5-VL 7B & 22.20 & 62.00 & 41.00 \\
    HuatuoGPT-V 7B & 21.40 & 65.00 & 46.58 \\
    Lingshu 7B & \underline{26.50} & 79.00 & 45.40 \\
    \rowcolor{green!12} Citrus-V 8B & \textbf{28.94} & \textbf{94.57} & \underline{51.07} \\
    \midrule
    \multicolumn{4}{c}{Open-Source Models (>10B)} \\
    \midrule
    MedPlib 14B & 0.07 & 0.04 & 21.05\\
    MedGemma 27B & 17.65 & 48.08 & \underline{48.73} \\
    Qwen2.5-VL 32B & 17.45 & 52.48 & 46.70 \\
    Lingshu 32B & \underline{25.29} & \underline{77.42} & 46.18 \\
    HealthGPT 14B & 21.29 & 68.24 & 47.82 \\
    HealthGPT 32B & 12.50 & 56.38 & 45.15 \\
    HuatuoGPT-V 34B & 23.97 & 66.07 & 45.51 \\
    \rowcolor{blue!12} Citrus-V 33B & \textbf{29.58} & \textbf{108.66} & \textbf{52.45} \\
    \bottomrule
  \end{tabular}%
  }
\end{table}

\subsubsection{Medical Image Detection and Segmentation}

Table \ref{tab:results:midsbwm} primarily presents a performance comparison of multimodal language models on detection and segmentation tasks. We evaluated segmentation performance across eight different image modalities using MedPlib 14B as a baseline, and our Citrus-V 8B model achieved the best segmentation results in all tasks. Since currently only MedPlib 14B and our model are capable of outputting image masks, we have not yet compared these results with other open-source models. Additionally, we assessed our model’s performance on detection tasks, where Citrus-V 8B demonstrated an 23.7\% improvement over the base model Qwen2.5-VL 7B.

To determine whether large multimodal language models can match the performance of domain-specific expert models in segmentation tasks, we conducted comparative tests against traditional expert models on several 2D image segmentation datasets, as shown in Table \ref{tab:results:midsbws}. Our Citrus-V 8B outperformed all expert models on 7 out of 8 test subsets and achieved performance metrics comparable to expert models on the remaining subset. These results indicate that question answering and segmentation tasks can be effectively integrated into a unified large multimodal language model, with performance that meets or even surpasses that of expert models.

\begin{table}[t!]
  \centering
  \caption{Performance of Models on Medical Image Detection and Segmentation Benchmarks Compared with MLLMs.
  The best results on each benchmark are highlighted in bold, and $^{\star}$ indicates zero-shot performance. For segmentation task, the evaluation metric is Dice. For detection task, the evaluation is Precision@0.5.}
  \label{tab:results:midsbwm}
  \resizebox{0.95\textwidth}{!}{%
  \begin{tabular}{lccccccccc}
    \toprule
    \multirow{2}{*}{\textbf{Model}} &
    \multicolumn{8}{c}{\textbf{MeCOVQA-G+}$_{seg}$} & \multicolumn{1}{c}{\textbf{MedSAM2}$_{det}$} \\
    \cmidrule(lr){2-10}
    & \textbf{DER} & \textbf{CT} & \textbf{PET} & \textbf{X-RAY} & \textbf{END} & \textbf{MR} & \textbf{US} & \textbf{FP} & - \\
    \midrule
    MedPlib 14B & 79.84 & 57.58 & 64.25 & 8.47$^{\star}$ & 44.35$^{\star}$ & 27.38$^{\star}$ & 34.22$^{\star}$ & 4.82$^{\star}$ & - \\
    Qwen2.5-VL 7B & - & - & - & - & - & - & - & - & 20.90 \\
    \rowcolor{green!12} Citrus-V 8B & \textbf{92.09} & \textbf{64.04} & \textbf{77.93} & \textbf{14.69} & \textbf{92.80} & \textbf{43.07} & \textbf{83.83} & \textbf{74.07} & \textbf{44.60} \\
    \bottomrule
  \end{tabular}%
  }
\end{table}

\begin{table}[t!]
  \centering
  \caption{Performance of Models on Medical Image Detection and Segmentation Benchmarks Compared with Specialized Models.
  The best results on each benchmark are highlighted in bold, and the scores with underline indicate the second best.}
  \label{tab:results:midsbws}
  \resizebox{\textwidth}{!}{%
  \begin{tabular}{lcccccccc}
    \toprule
    \textbf{Model} &
    \multicolumn{8}{c}{\textbf{MedSegBench}} \\
    \cmidrule(lr){2-9}
    & \textbf{Isic2016} & \textbf{Kvasir} & \textbf{Idrib} & \textbf{CovidQUEx} & \textbf{Promise12} & \textbf{MosMedPlus} & \textbf{UltrasoundNerve} & \textbf{Tnbcnuclei} \\
    \midrule
    Unet-MIT & 89.10 & 56.90 & 5.30 & - & - & 76.10 & - & 75.90 \\
    Unet-EfficientNet & \underline{90.30} & \underline{81.20} & 7.80 & 74.40 & 89.20 & 78.10 & \underline{78.70} & 73.80 \\
    Unet-MobileNetV2 & 89.10 & 75.40 & 9.20 & 74.20 & 89.60 & 78.50 & 77.20 & 76.20 \\
    Unet-DenseNet121 & 89.30 & 79.40 & 8.90 & \underline{75.60} & \underline{90.00} & \textbf{79.10} & 78.60 & \underline{78.80} \\
    Unet-ResNet18 & 87.80 & 73.90 & \underline{10.00} & 74.00 & 89.50 & 78.00 & 78.20 & 77.90 \\
    Unet-ResNet50 & 88.70 & 69.80 & 9.00 & 73.40 & 88.80 & \underline{79.00} & 77.60 & 78.50 \\
    \rowcolor{green!12} Citrus-V 8B & \textbf{93.16} & \textbf{90.91} & \textbf{47.10} & \textbf{77.84} & \textbf{90.81} & 78.17 & \textbf{80.15} & \textbf{82.93} \\
    \bottomrule
  \end{tabular}%
  }
\end{table}

\subsection{Ablation Studies}
\label{sec:ablation}

\subsubsection{Training Stages}

First, we evaluated the impact of the alignment stages on model capabilities using a 7B-scale model, focusing specifically on its performance in medical visual question answering tasks. Detailed results are presented in Table \ref{tab:results:asdtsmvqab}. The results show that, after applying the first two alignment stages, the model’s average score across various task metrics increased by 4.71\% compared to the baseline model. This demonstrates that fine-tuning the visual encoder during the alignment stage is essential for enhancing the model’s understanding of medical imaging modalities.

\begin{table}[t!]
  \centering
  \caption{Ablation Studies with Different Training Stages on Medical Visual Question Answering Benchmarks. We mainly focus on the impact of the first 2 alignment training stages.}
  \label{tab:results:asdtsmvqab}
  \resizebox{0.9\textwidth}{!}{%
  \begin{tabular}{lcccccc}
    \toprule
    \textbf{Model} & \textbf{VQA-RAD} & \textbf{MedXpertQA} & \textbf{SLAKE} & \textbf{PATH-VQA} & \textbf{PMC-VQA} & \textbf{Avg.} \\
    \midrule
    Qwen2.5-VL 7B & 66.30 & 20.75 & 67.86 & 42.30 & 50.86 & 49.61 \\
    \quad + \colorbox{yellow!12}{stage 3} & 63.60 & 19.95 & 70.77 & 46.55 & 55.63 & 51.30\imp{1.69} \\
    \quad + \colorbox{yellow!12}{stage 1\&2\&3} & 67.18 & 23.45 & 75.36 & 47.91 & 57.71 & 54.32\imp{4.71} \\
    \bottomrule
  \end{tabular}%
  }
\end{table}

\begin{table}[t!]
  \centering
  \caption{Ablation Studies with Different Training Stages on Medical Images Detection and Segmentation Benchmarks. We mainly focus on the impact of 3rd and 4th training stages.}
  \label{tab:results:asdtsmidsb}
  \resizebox{0.95\textwidth}{!}{%
  \begin{tabular}{lccccccccc}
    \toprule
    \multirow{2}{*}{\textbf{Model}} &
    \multicolumn{8}{c}{\textbf{MeCOVQA-G+}$_{seg}$} & \multicolumn{1}{c}{\textbf{Avg.}} \\
    \cmidrule(lr){2-10}
    & \textbf{DER} & \textbf{CT} & \textbf{PET} & \textbf{X-RAY} & \textbf{END} & \textbf{MR} & \textbf{US} & \textbf{FP} & - \\
    \midrule
    Qwen2.5-VL 7B & - & - & - & - & - & - & - & - & - \\
    \quad + \colorbox{yellow!12}{stage 1\&2\&3} & 88.22 & 59.46 & 69.10 & 27.20 & 90.48 & 36.14 & 79.72 & 95.47 & 66.58 \\
    \quad + \colorbox{yellow!12}{stage 1\&2\&3\&4} & 92.09 & 64.04 & 77.93 & 14.69 & 92.80 & 43.07 & 83.83 & 74.07 & 67.82\imp{1.24} \\
    \bottomrule
  \end{tabular}%
  }
\end{table}

Second, we conducted ablation experiments to assess the impact of the fourth alignment stage on segmentation tasks. As shown in Table \ref{tab:results:asdtsmidsb}, training with the fourth stage further improved the model’s segmentation accuracy by 1.34\%, while maintaining its performance on other understanding tasks. This result confirms the positive effect of fine-tuning during the fourth stage.

\subsubsection{Data Types}

We conducted detailed experiments to evaluate the impact of textual reasoning data and visual grounding data on model capabilities. Specifically, textual reasoning data was primarily used to assess the model’s performance in question-answering tasks, while visual grounding data was mainly utilized to evaluate the model’s ability to ground anatomical structures during multimodal chain-of-thought reasoning.

\paragraph{Medical Texual Reasoning Data}

From the experimental results presented in Tables \ref{tab:results:mvqab} and \ref{tab:results:mtqab}, we observe that commercial models tend to exhibit weaker perceptual capabilities for medical images, as demonstrated on test sets such as VQA-RAD, SLAKE, PATH-VQA, and PMC-VQA. However, they possess exceptionally strong reasoning abilities, as evidenced by their performance on test sets like MedXpertQA and Medbullets. Therefore, this section primarily investigates strategies to enhance the reasoning capabilities of multimodal language models, with a focus on analyzing the impact of two types of data: the medical text reasoning dataset Citrus S3 and the open-source mathematical reasoning dataset AQUA.

In addition to these two datasets, to balance the data distribution across various tasks and improve model performance on downstream tasks, we constructed a balanced dataset of approximately one million samples through random sampling, named CitrusV-1M. This dataset encompasses a diverse array of text-based and multimodal question answering and report generation samples.

During the third stage, we conducted three groups of experiments, sequentially incorporating Citrus-1M, AQUA, and Citrus S3 datasets to evaluate the model’s performance on text and visual question answering tasks involving reasoning. The results of the ablation experiments are presented in Table x. Our observations are as follows. First, after incorporating only Citrus-1M data, the model’s performance on medical reasoning datasets improved by 5.78\% over the base model. Second, further addition of AQUA data, although not medically related, provided a positive impact not only on medical text reasoning evaluation sets but also on medical visual reasoning sets such as MedXpertQA-MM, with an average improvement of 1.96\%. Third, by further increasing the amount of Citrus S3 data, the model’s average performance on reasoning-related evaluation sets was significantly enhanced, with an improvement of approximately 1.35\%. Notably, even when only pure text reasoning data was added, the model showed improvements not only on text reasoning evaluation sets but also on visual reasoning sets such as MedXpertQA-MM.

The ablation experiment results demonstrate that augmenting the model with both medical pure text reasoning datasets and non-medical reasoning datasets can effectively enhance its reasoning ability in the medical question answering domain. This also indirectly illustrates that, despite commercial models not being specifically optimized for the medical domain, their robust general reasoning capabilities enable them to achieve strong performance on medical evaluation sets.

\begin{table}[t!]
  \centering
  \caption{Ablation Studies with Medical Textual Reasoning Data on Medical Reasoning Question Answering Benchmarks.}
  \label{tab:results:asmtrdmrqab}
  \resizebox{0.9\textwidth}{!}{%
  \begin{tabular}{lccccccc}
    \toprule
    \multirow{2}{*}{\textbf{Model}} & \multicolumn{2}{c}{\textbf{MedXpertQA}} & \multicolumn{2}{c}{\textbf{Medbullets}} & \textbf{MedQA} & \textbf{PubMedQA} &
    \multirow{2}{*}{\textbf{Avg.}} \\
    \cmidrule(lr){2-7}
    & \textbf{MM} & \textbf{Text} & \textbf{Op4} & \textbf{Op5} & \textbf{-} & \textbf{-} & \\
    \midrule
    Qwen2.5-VL 7B & 20.75 & 12.40 & 47.08 & 36.69 & 57.50 & 75.8 & 41.70 \\
    \quad + \colorbox{yellow!12}{Citrus-1M} & 24.00 & 15.10 & 58.12 & 48.70 & 61.74 & 77.2 & 47.48\imp{5.78} \\
    \quad + \colorbox{yellow!12}{Citrus-1M + AQUA} & 25.15 & 16.16 & 61.04 & 51.95 & 64.10 & 78.20 & 49.44\imp{7.74} \\
    \quad + \colorbox{yellow!12}{Citrus-1M + AQUA + Citrus S3} & 24.8 & 16.49 & 59.42 & 59.74 & 67.71 & 76.6 & 50.79\imp{9.09} \\
    \bottomrule
  \end{tabular}%
  }
\end{table}

\section{Conclusions}
In this work, we introduced Citrus-V, a multimodal medical foundation model that unifies pixel-level image grounding with clinical reasoning. By integrating detection, segmentation, and multimodal chain-of-thought reasoning within a single framework, Citrus-V delivers a cohesive pipeline from lesion localization to structured reporting and diagnostic inference. Our novel multimodal training paradigm, combined with the release of a curated open-source data suite, further enhances the development and reproducibility of general-purpose medical foundation models.

Extensive experiments demonstrate that Citrus-V outperforms prior open-source medical foundation models and expert-level imaging systems, setting new benchmarks across both visual and multimodal reasoning tasks. 
We achieved significant improvements over other open-source models of the same size across five major categories of evaluation sets. Notably, in medical document understanding tasks, our model is able to outperform commercial models. In medical image segmentation tasks, our model’s performance matches or even surpasses that of expert models. For report generation tasks, we introduced multimodal chain-of-thought reasoning to realistically emulate the diagnostic process of physicians, resulting in substantial performance gains on the test sets.

Looking ahead, Citrus-V represents a significant step toward clinically reliable foundation models, offering the potential to support diverse workflows such as precise lesion quantification, automated radiology reporting, and reliable second opinions. By bridging visual grounding with reasoning, Citrus-V paves the way for the next generation of medical AI systems with greater generalizability, interpretability, and clinical utility.

\newpage

{
    \bibliographystyle{unsrt}
    \bibliography{bibs/references}
}

\clearpage
\appendix

\section{Case Studies}\label{app:cases}

\noindent\makebox[\linewidth][c]{
  \includegraphics[page=1,height=.47\textheight,keepaspectratio,trim=5cm 0 5cm 0,clip]{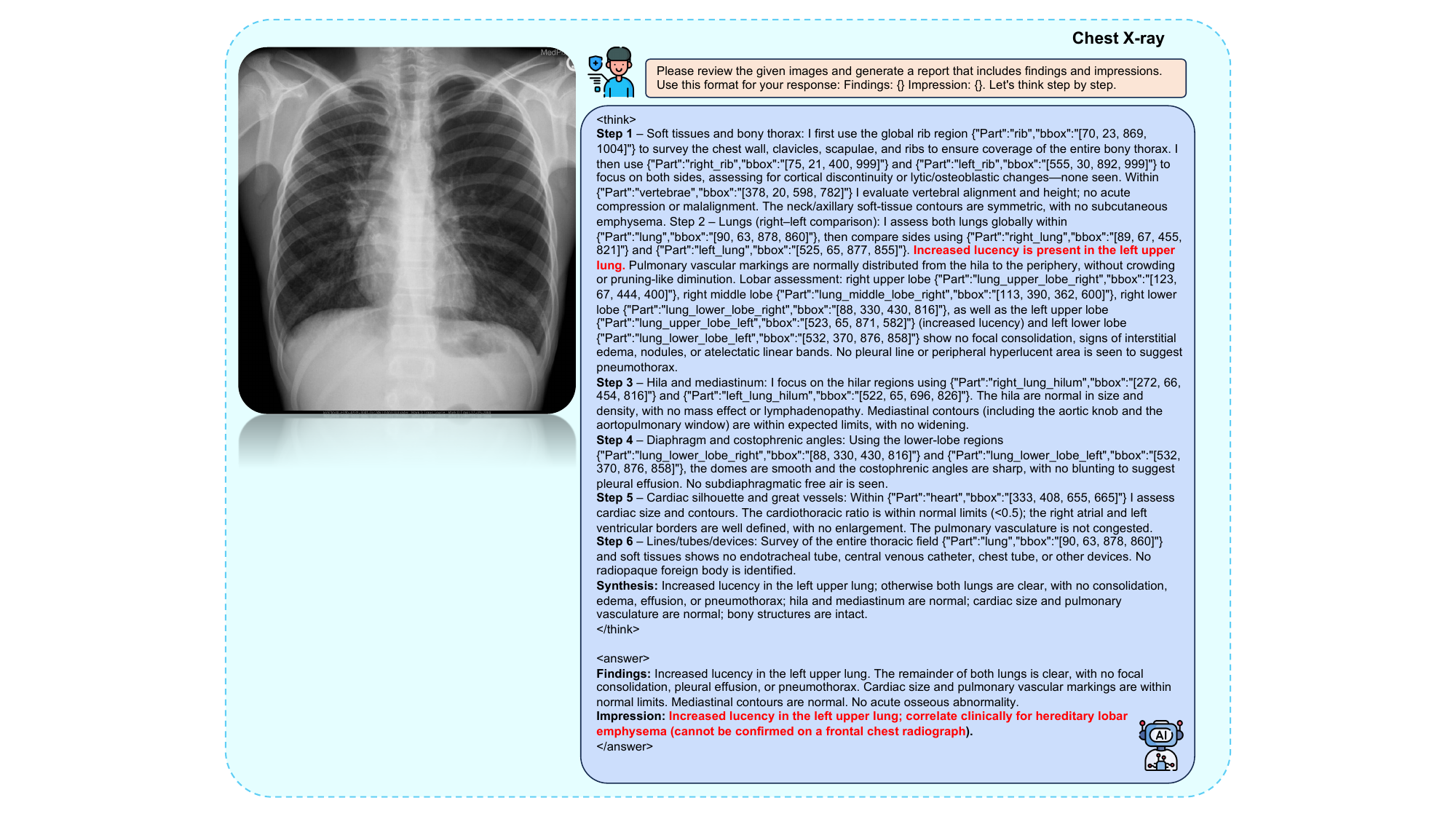}
}
\noindent\makebox[\linewidth][c]{
  \includegraphics[page=2,height=.47\textheight,keepaspectratio,trim=5cm 0 5cm 0,clip]{figs/casestudy.pdf}
}

\foreach \p in {3,...,8}{
  \begin{figure}[t!]
    \centering
    \includegraphics[page=\p,width=\linewidth,trim=5cm 0 5cm 0,clip]{figs/casestudy.pdf}
  \end{figure}
}

\begin{figure}[t!]
  \centering
  \includegraphics[page=9,width=\linewidth,trim=3.5cm 0 3.3cm 0,clip]{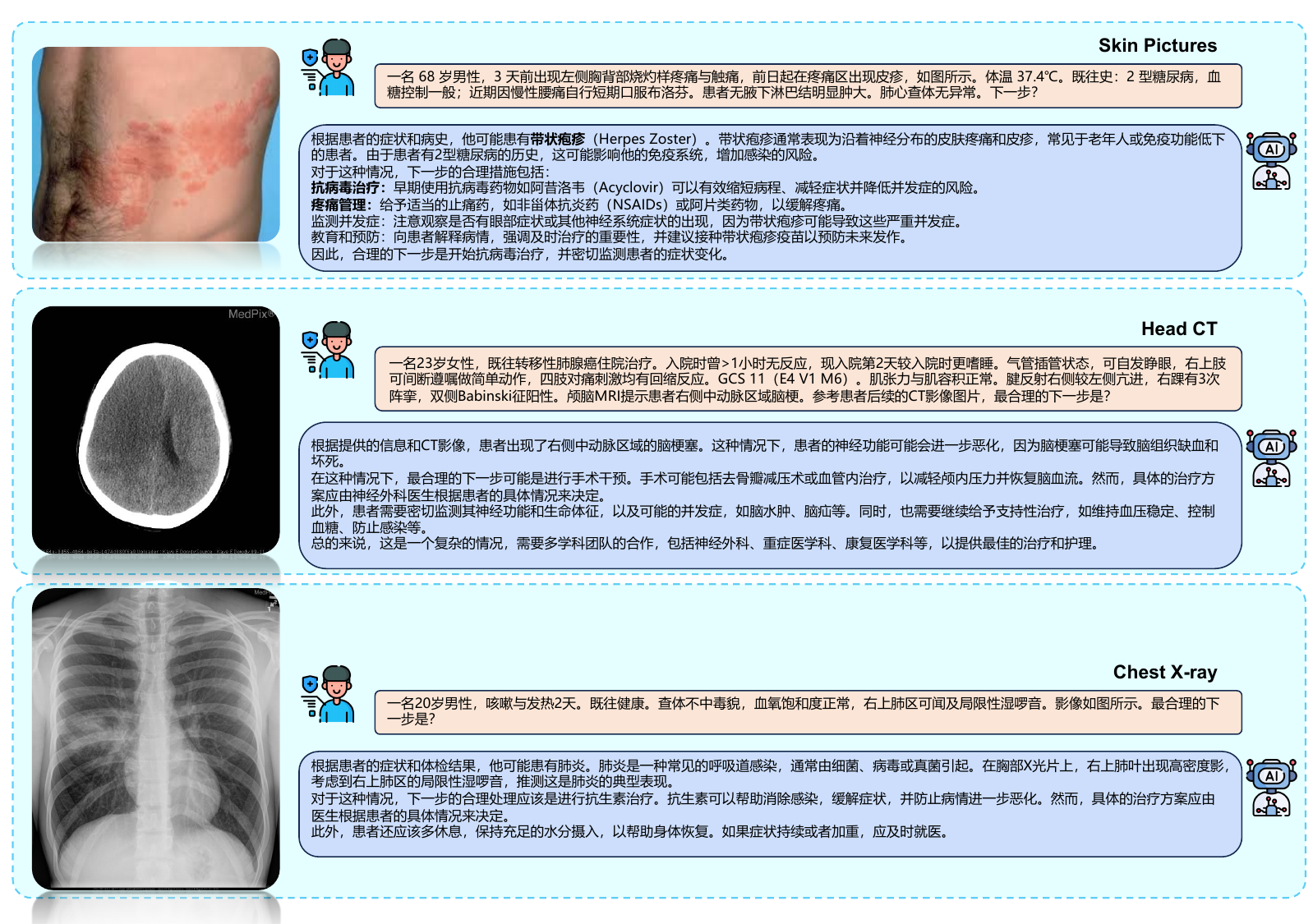}
\end{figure}
\begin{figure}[t!]
  \centering
  \vspace{-25ex}
  \includegraphics[page=10,width=\linewidth,trim=3.5cm 0 3.3cm 0,clip]{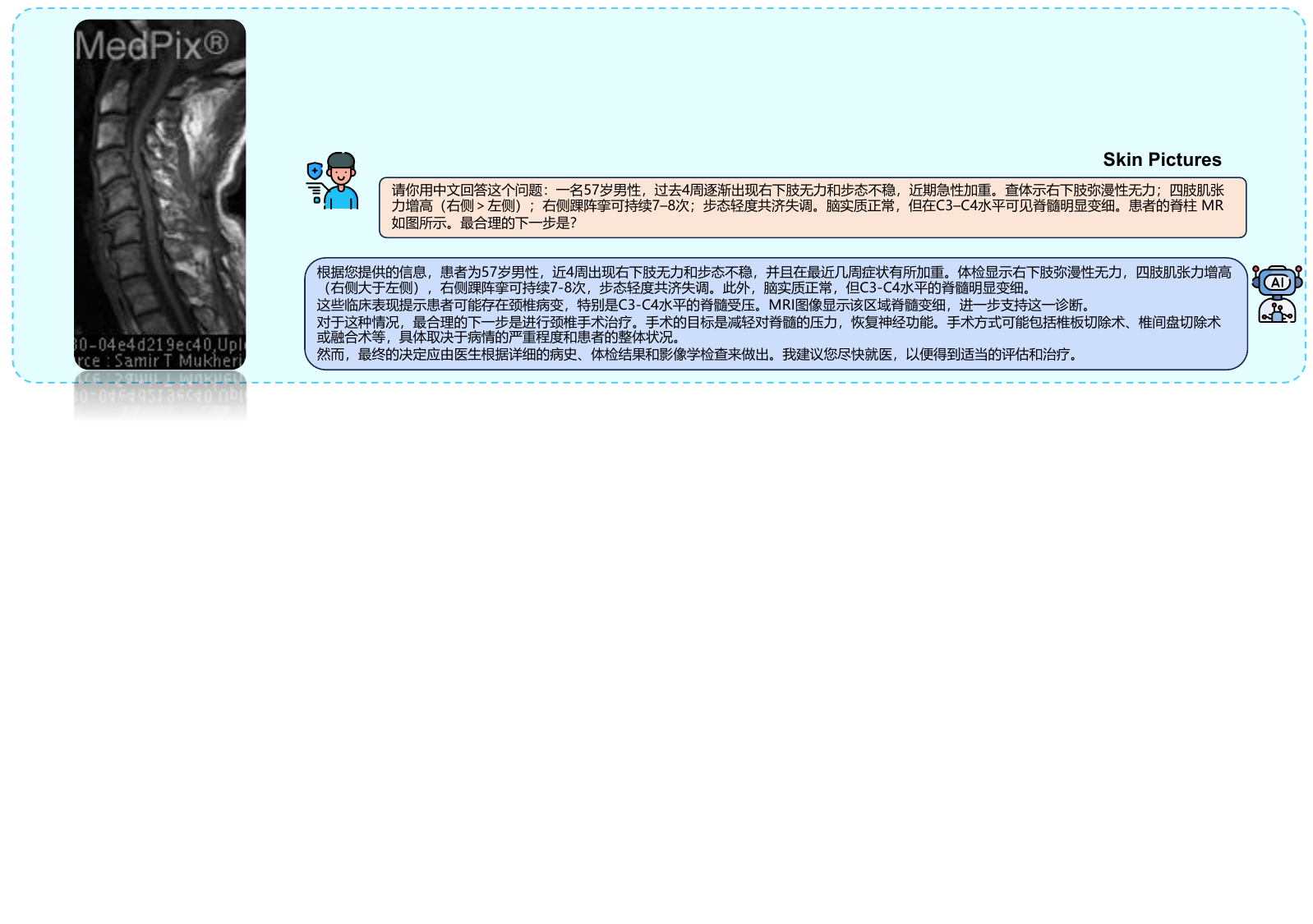}
\end{figure}



\end{document}